\documentclass[letterpaper]{article} % DO NOT CHANGE THIS
\usepackage[draft]{aaai2026}  % DO NOT CHANGE THIS
\usepackage{times}  % DO NOT CHANGE THIS
\usepackage{helvet}  % DO NOT CHANGE THIS
\usepackage{courier}  % DO NOT CHANGE THIS
\usepackage[hyphens]{url}  % DO NOT CHANGE THIS
\usepackage{graphicx} % DO NOT CHANGE THIS
\usepackage{natbib}  % DO NOT CHANGE THIS AND DO NOT ADD ANY OPTIONS TO IT
\usepackage{caption} % DO NOT CHANGE THIS AND DO NOT ADD ANY OPTIONS TO IT
\usepackage{booktabs}
\usepackage{subcaption}
\usepackage{float}
\usepackage{booktabs}
\usepackage{booktabs}
\usepackage{multirow}
\usepackage{makecell}
\usepackage[table]{xcolor}
\definecolor{OursRow}{HTML}{EEF4FB}
\usepackage{algorithm}
\usepackage{algorithmic}
\usepackage{booktabs}
\usepackage{array}
\usepackage[table]{xcolor}
\usepackage{booktabs}
\usepackage{tabularx}
\usepackage{array}
\usepackage{amsmath}
\usepackage{listings}
\usepackage{xcolor}
\lstdefinestyle{prompt}{%
  basicstyle=\ttfamily\scriptsize,
  breaklines=true,
  columns=fullflexible,
  frame=single,
  rulecolor=\color{gray!45},
  backgroundcolor=\color{gray!6},
  framexleftmargin=3pt,
  xleftmargin=3pt,
  breakindent=0pt,
  keepspaces=true,
}

\definecolor{MetricArrow}{HTML}{4F78A8}

\usepackage{enumitem}
\usepackage{xcolor}
\usepackage{listings}
\usepackage{graphicx}
\definecolor{OSWorldBlue}{HTML}{EAF2F8}
\definecolor{OSWorldBlueText}{HTML}{1F4E79}
\definecolor{CUAOrange}{HTML}{FDF0E6}
\definecolor{CUAOrangeText}{HTML}{9A4D00}
\newcommand{\tool}[1]{\texttt{\small #1}\hspace{0.3em}\allowbreak}
\newcommand{\famlabel}[1]{\textsc{\footnotesize #1}}
\newlist{toollist}{description}{1}
\setlist[toollist]{leftmargin=6.5em, style=sameline, font=\normalfont,
                     itemsep=3pt, parsep=0pt, topsep=3pt, before=\raggedright}

\usepackage[scaled=0.85]{beramono}
\usepackage{newfloat}
\usepackage{listings}
\DeclareCaptionStyle{ruled}{labelfont=normalfont,labelsep=colon,strut=off} % DO NOT CHANGE THIS
\floatstyle{ruled}
\newfloat{listing}{tb}{lst}{}
\floatname{listing}{Listing}
\title{%
\begin{tabular}[c]{@{}c@{}c@{}}
\raisebox{-0.38\height}{\includegraphics[height=2.4em]{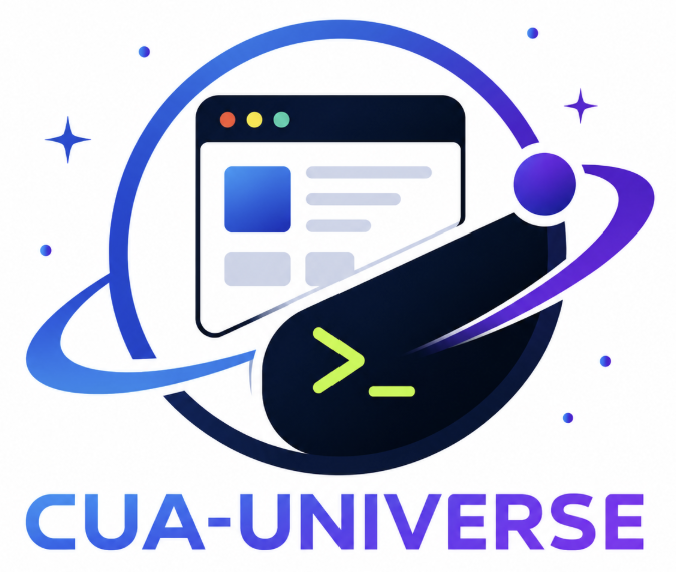}}
&
\begin{tabular}[c]{c}
CUA-Universe: A Scalable and Dynamic Environment\\
for Hybrid GUI+CLI Agents
\end{tabular}
\end{tabular}%
}

\author{
Haoting Shi\textsuperscript{\rm 1}\equalcontrib,
Wenhao Wang\textsuperscript{\rm 2}\equalcontrib\textsuperscript{\rm \textdagger},
Weicheng Fang\textsuperscript{\rm 2},
Yaozhong Liang\textsuperscript{\rm 2},
Tian Jin\textsuperscript{\rm 1},\\
Pengxiang Zhao\textsuperscript{\rm 2},
Guangyi Liu\textsuperscript{\rm 2},
Siheng Chen\textsuperscript{\rm 1}\textsuperscript{\rm \textdagger},
Yanfeng Wang\textsuperscript{\rm 1}
}

\affiliations{
\textsuperscript{\rm 1}Shanghai Jiao Tong University \qquad
\textsuperscript{\rm 2}Zhejiang University\\
}
\definecolor{OursRow}{HTML}{EEF4FB}

\begin{document}

\maketitle

\begingroup
\renewcommand{\thefootnote}{\fnsymbol{footnote}}
\footnotetext[2]{Corresponding authors.}
\endgroup

\begin{abstract}
Computer-use agents have advanced on benchmarks like OSWorld and AndroidWorld, but still act mostly through the GUI, often producing inefficient trajectories. Real-world computer work is hybrid, combining visual-state inspection with precise, high-throughput command-line operations, so capable agents must coordinate both modalities over shared application state. Yet scalable hybrid environments remain scarce because supporting both GUI and CLI over real applications typically requires substantial manual engineering for each application. Existing agents also struggle to use the two interfaces complementarily: CLI-native agents lack visual perception for tasks involving interface state or layout, while GUI-native agents are inefficient for operations better executed through commands.

We introduce \textbf{CUA-Universe}, a scalable environment-to-data pipeline that turns real desktop software into hybrid GUI+CLI environments. \emph{App-Forge} adapts applications into reproducible VMs and command-line surfaces it discovers, wraps, or generates, scaling to 16 applications across diverse domains; \emph{Task-Weave} synthesizes diverse hybrid tasks of controllable difficulty from reusable operations over seed files, turning each environment into a continuous task source; and \emph{Path-Steer} rolls agents out along efficient hybrid paths and harvests verified trajectories for post-training. Training on this data shifts behavior from inefficient GUI interaction and brittle CLI scripting toward effective GUI+CLI orchestration, improving both success and efficiency for our 9B model on CUA-Verse (Score $+39.3$ pts; $-37\%$ steps, $-60\%$ tokens), OSWorld (SR $+16.8$ pts; $-57\%$ steps, $-44\%$ tokens), and OSWorld-MCP (Score $+7.84$ pts; $-27\%$ steps, $-30\%$ tokens). By converting real desktop software into hybrid environments and reusable training data, CUA-Universe provides a scalable path toward more capable and efficient computer-use agents. We will release our code and data.
\end{abstract}

\begin{figure*}[t]
    \centering
 \includegraphics[width=1\linewidth]{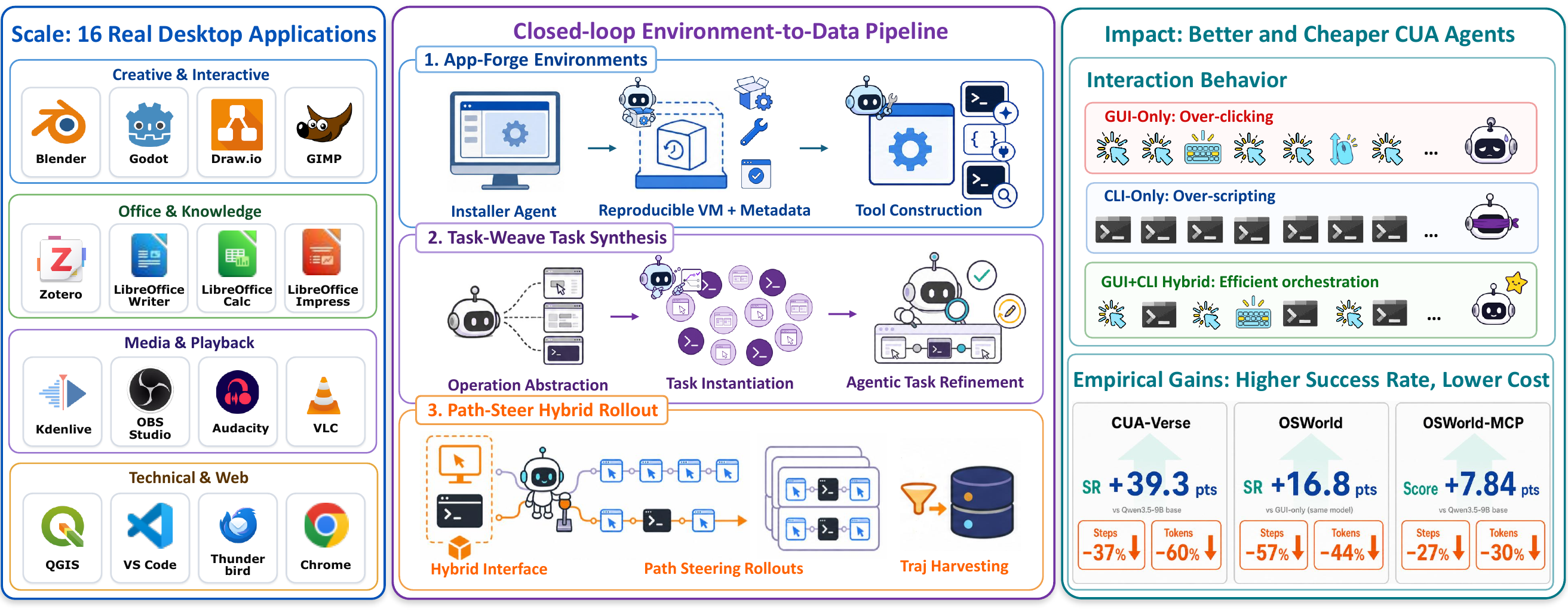}
\caption{\textbf{Overview of CUA-Universe.} \textbf{(1) Scale:} 16 real desktop applications spanning creative \& interactive, office \& knowledge, media \& playback, and technical \& web domains, each exposed through both a GUI and application-specific CLI interfaces. \textbf{(2) Closed-loop environment-to-data pipeline:} \emph{App-Forge} adapts an application into a reproducible VM and the command-line surface it discovers, wraps, or generates (installer agent~$\rightarrow$~reproducible VM~$+$~metadata~$\rightarrow$~tool construction); \emph{Task-Weave} synthesizes diverse hybrid GUI+CLI tasks at scale over seed files (operation abstraction~$\rightarrow$~task instantiation~$\rightarrow$~agentic refinement); and \emph{Path-Steer} rolls agents out onto efficient hybrid paths and harvests high-quality, directly reusable trajectories for post-training. \textbf{(3) Impact:} training on this data shifts interaction behavior from inefficient GUI interaction and brittle CLI scripting toward effective GUI+CLI orchestration, producing more capable and efficient agents with consistent improvements across diverse benchmarks, including CUA-Verse (Score $+39.3$ pts; $-37\%$ steps, $-60\%$ tokens), OSWorld (SR $+16.8$ pts; $-57\%$ steps, $-44\%$ tokens), and OSWorld-MCP (Score $+7.84$ pts; $-27\%$ steps, $-30\%$ tokens).}

    \label{fig:pipeline}
\end{figure*}

\section{Introduction}
Computer-use agents (CUAs) have advanced rapidly, completing a broad range of real desktop and mobile tasks on benchmarks such as OSWorld and AndroidWorld~\cite{xie2024osworld,rawles2025androidworld}. Yet leading agents still interact predominantly through the graphical user interface (GUI), resulting in trajectories that are often unnecessarily long~\cite{abhyankar2025osworld} and increasingly brittle on extended workflows~\cite{yuan2026osworld2}. Real computer use, however, is inherently multi-modal: users rely on the GUI for visually grounded interaction, while using command-line tools, scripts, or APIs for precise and high-throughput operations. Capable CUAs should therefore operate in \emph{hybrid GUI+CLI environments}, dynamically choosing the interface best suited to each subtask while maintaining a shared application state across modalities~\cite{song2025coact,yang2025ultracua,jia2025osworld}.

Realizing such hybrid computer-use agents faces two coupled bottlenecks: \emph{scalable real-world environments} and \emph{cross-modality orchestration}. On the environment side, GUI-centric environments are costly to build and often require substantial manual engineering, while CLI-centric environments are easier to automate but lack access to visual layout and interface state. Building hybrid environments that support both modalities over the same application state therefore remains expensive and difficult to scale across applications. On the agent side, existing agents are typically optimized for one modality and struggle to use the two interfaces complementarily. CLI-native agents often lack visual perception and therefore resort to brittle scripts for tasks that depend on interface state or visual layout, while GUI-native agents can be inefficient for batch operations that a single command could complete. The deeper challenge is therefore \emph{orchestration}: deciding when to switch interfaces and carrying task state across them, for example, locating a target through the GUI and then processing it through the CLI. 

To address these bottlenecks, we introduce \textbf{CUA-Universe}, a real-software environment framework for synthesizing, evaluating, and training hybrid GUI+CLI computer-use agents. CUA-Universe organizes real software into a scalable environment-to-data pipeline with three components. \textbf{App-Forge} adapts a desktop application into a reproducible VM and exposes it through a command-line surface it discovers, wraps, or generates. Driven by a coding agent rather than per-application manual engineering, it scales CUA-Universe to 16 desktop applications across diverse domains. \textbf{Task-Weave} synthesizes GUI+CLI hybrid tasks of controllable difficulty from applications, tools, and seed states, turning each environment into a continuous source of tasks. \textbf{Path-Steer} guides agents toward efficient hybrid execution paths, using the CLI for batch and precise operations and the GUI for visually grounded ones, and harvests high-quality trajectories for post-training. Together, these components turn real software into scalable sources of hybrid tasks and training data, enabling agents to learn more effective cross-modality orchestration between GUI and CLI.

Our evaluation spans three complementary axes. First, to directly evaluate a model's ability to orchestrate GUI and CLI actions over a shared application state, we construct \textbf{CUA-Verse}, a held-out benchmark of 160 hybrid tasks built on the eight desktop applications introduced by CUA-Universe, with evaluation tasks disjoint from the training data. On CUA-Verse, our model improves success by roughly $3\times$ over its identical base while using $60\%$ fewer tokens, achieving the best open-source result. Second, we test whether this capability transfers beyond our benchmark to OSWorld, where our model gains $+16.8$ success-rate points over GUI-only execution while using fewer steps and tokens. Finally, on OSWorld-MCP~\cite{jia2025osworld}, our model generalizes to a different tool-invocation interface unseen during training, improving Score by $+7.84$ points over its base. Together, these results show that CUA-Universe improves not only performance within the environments it constructs, but also hybrid interaction skills that transfer across tasks, benchmarks, and tool interfaces.

We summarize our contributions as follows:
\begin{itemize}
   \item \textbf{A scalable environment-to-data pipeline.} CUA-Universe turns real desktop software into hybrid GUI+CLI agent environments through three components, App-Forge, Task-Weave, and Path-Steer, which cover application adaptation, task synthesis, and trajectory generation. Agent-driven construction allows the framework to scale efficiently to 16 real applications across diverse domains.
   
    \item \textbf{Scalable hybrid task synthesis and efficient trajectory generation.} The framework discovers, wraps, or generates application-specific command-line interfaces, synthesizes GUI+CLI tasks of controllable difficulty from reusable operations and seed states, and steers agents toward efficient hybrid execution paths. This turns each environment into a continuous source of tasks and training trajectories while encouraging more effective cross-modality orchestration.

    \item \textbf{A benchmark for hybrid GUI+CLI orchestration.} \textbf{CUA-Verse} provides 160 held-out hybrid tasks across eight desktop applications introduced by CUA-Universe, directly evaluating an agent's ability to coordinate GUI and CLI actions over a shared application state.

    \item \textbf{Strong performance and transfer.} Our 9B model trained on CUA-Universe data achieves roughly $3\times$ the success of its base model on CUA-Verse while using $60\%$ fewer tokens. The learned hybrid interaction skills further transfer to OSWorld with a $+16.8$ point gain over GUI-only execution and generalize to the unseen tool interface of OSWorld-MCP.

\end{itemize}

\section{Related Work}
\subsection{Environment Synthesis for Computer-Use Agents}
A growing body of work automatically synthesizes environments and trajectories to avoid the cost of hand-curated benchmarks~\cite{zhou2024webarena, deng2023mind2web, xie2024osworld, rawles2025androidworld}. InfiniteWeb~\cite{zhang2026infiniteweb}, GUI-Genesis~\cite{cao2026gui}, and AutoWebWorld~\cite{wu2026autowebworld} generate functional web environments for post-training, but their targets are artificial web pages confined to the GUI modality, leaving a \emph{sim-to-real gap} on actual software. Scaling further, CUA-Gym~\cite{wang2026cua} co-generates environments, tasks, and verifiable rewards across desktop and mock web applications for RLVR, and Gym-Anything~\cite{aggarwal2026gym} turns arbitrary applications into agent environments and distills successful trajectories into a model. These environments, however, are driven purely through the GUI, so the trajectories they yield are inherently single-modality and cannot exhibit \emph{when} to leave the GUI. In contrast, CUA-Universe builds each application into a hybrid GUI+CLI environment by exposing it through both the GUI and application-specific CLI tools. The synthesized tasks therefore require coordinating the two modalities over a shared application state, while the harvested trajectories carry efficiency and orchestration signals that a GUI-only pipeline structurally cannot provide.

\subsection{Single-Modality GUI and CLI Agents}
Computer-use agents have progressed along two separate lines. On the GUI side, agents are increasingly evaluated across diverse desktop, mobile, and heterogeneous platform settings, including OSWorld~\cite{xie2024osworld}, AndroidWorld~\cite{rawles2025androidworld}, and FedGUI~\cite{wang-etal-2025-fedmabench, wang2026fedgui}, but may remain inefficient, taking far longer trajectories than necessary~\cite{abhyankar2025osworld} and degrading on long-horizon, cross-application workflows~\cite{yuan2026osworld2}. On the CLI side, the terminal has become a first-class target: Terminal-Bench~\cite{merrill2026terminal} and TerminalWorld~\cite{chu2026terminalworld} provide hard command-line tasks, CLI-Universe~\cite{hua2026cli} synthesizes verifiable terminal tasks, and further work scales terminal training environments and recipes~\cite{cheng2026terminal, ivison2026tmax}. Yet competence in one modality does not transfer to the other: terminal agents cannot handle operations that depend on visual state, while GUI agents fall back to slow, element-by-element manipulation for batch operations.  CUA-Universe instead trains on GUI+CLI hybrid tasks over a shared application state, teaching agents \emph{when} to switch modalities, a form of coordination that neither line develops in isolation and that improves both success rate and execution efficiency.

\subsection{Toward Hybrid GUI+CLI Agents}
A recent trend combines visual actions with CLI tool calls rather than relying on either alone: CoAct-1~\cite{song2025coact} pairs a GUI operator with a coding agent, and UltraCUA~\cite{yang2025ultracua} trains a hybrid-action model whose tools are mined from generic documentation and code repositories over a fixed task set. Hybrid benchmarks further quantify the GUI+CLI efficiency trade-off~\cite{li2026weavebench, zhou2026gui, fu2026macagentbench}. Unlike these existing works, which train a hybrid agent over a \emph{fixed} set of tools and tasks, CUA-Universe contributes a new way to \emph{construct environments and synthesize data}. Rather than mining a generic tool set, it grounds the tool layer in \emph{each real desktop application}. Inspired by CLI-Anything~\cite{yang2026cli}, we generate agent-native CLIs when needed while also exposing applications' native command-line tools, turning each environment into a continuous source of tasks and trajectories. On top of this, \emph{Path-Steer} steers rollouts onto efficient hybrid paths, raising success rates and yielding higher-quality trajectories for computer-use agent post-training.

\section{Method}

\subsection{Overview and Problem Formulation}
\label{sec:method-overview}

\paragraph{Overview.}
As illustrated in Figure~\ref{fig:pipeline}, CUA-Universe turns real desktop applications into hybrid GUI+CLI environments and further converts them into scalable sources of tasks and training trajectories. The framework consists of three components. \emph{App-Forge} (\S\ref{sec:env}) scales environment construction by adapting applications into reproducible hybrid environments with programmatic tool surfaces. \emph{Task-Weave} (\S\ref{sec:task-synth}) scales task generation by composing diverse hybrid tasks from reusable operations grounded in real application states. \emph{Path-Steer} (\S\ref{sec:rollout}) scales trajectory collection by steering agents toward efficient GUI+CLI execution paths and harvesting rollouts for post-training.

\paragraph{Problem formulation.}

We model a hybrid task as a POMDP: at step $t$ the agent observes $o_t$ (a screenshot plus optional textual CLI returns) and emits $a_t\in\mathcal{A}_{\text{gui}}\cup\mathcal{A}_{\text{cli}}$, where both action spaces read and modify over a shared persistent application state $s_t$. A task is a tuple $\tau=(\text{instr},s_0,V)$ of an instruction, a seed initial state, and a verifier $V(\zeta)\in[0,1]$ scoring a trajectory $\zeta$ (a VLM judge); the pipeline synthesizes such tasks at scale and collects trajectories that solve them along efficient hybrid paths.

\subsection{App-Forge: Scalable Agentic Environment Construction}
\label{sec:env}

Scaling hybrid environments across real desktop applications faces two application-specific bottlenecks: \emph{environment adaptation}, since applications differ substantially in installation, configuration, and runtime dependencies; and \emph{tool construction}, since agents need usable interfaces across heterogeneous applications. App-Forge addresses both through a scalable agentic construction pipeline that produces reproducible application environments together with CLI surfaces aligned with their GUIs.

\paragraph{Application adaptation.}
The first bottleneck is reproducibly adapting diverse desktop software, whose installation procedures, dependencies, and launch configurations vary substantially across applications. An installer agent, guided by an installation skill, operates a persistent VM harness to install and configure the target application and its supporting command-line utilities, interactively diagnosing failures until successful launch is verified. The resulting setup is distilled into a reproducible configuration, while lightweight introspection extracts application metadata for later task grounding and verification. The same adaptation workflow is reused across applications, allowing CUA-Universe to scale to 16 desktop applications across diverse domains. Eight are inherited from OSWorld but were originally GUI-only, for which App-Forge adds a shared-state CLI layer, while the other eight are introduced by CUA-Universe (Appendix~\ref{app:apps}).

\paragraph{Tool construction.}
\label{sec:tool}
The second bottleneck is constructing an expressive CLI surface across heterogeneous applications. We draw on three sources: native command-line tools we \emph{discover} (e.g., \texttt{blender -{}-python-expr}, \texttt{cvlc}), scripting APIs we \emph{wrap} (e.g., \texttt{bpy}, LibreOffice UNO, and GIMP Script-Fu), and agent-native CLIs we \emph{generate} when existing interfaces are insufficient, inspired by CLI-Anything. This layered design supports diverse automation surfaces without requiring a manually designed CLI for every application. GUI and CLI operate over the same project state, with lightweight adapters resynchronizing stale GUI views after external CLI edits. Appendix~\ref{app:tools} lists the complete tool inventory for all applications.

\subsection{Task-Weave: Compositional Hybrid Task Synthesis}
\label{sec:task-synth}
Scaling task generation across real applications requires synthesizing tasks that reflect executable capabilities, cover diverse compositions, and remain feasible in the live environment. Task-Weave addresses these requirements in three stages: it abstracts reusable operations from agent exploration, composes them into diverse seed-conditioned tasks, and validates the resulting tasks through real execution.

\paragraph{Operation abstraction.}
We first build a reusable operation pool that captures what can be reliably performed in each application. For a family of exploration seeds, we run parallel exploration agents with \emph{diverse goals}, each targeting a different facet of the application, such as structure and visibility, appearance, or export and organization. Each agent interacts with the GUI while recording screenshots and actions. We slide a window over each trajectory and prompt an LLM to abstract the interaction into a high-level reusable \emph{operation}, such as \texttt{export\_scene\_to\_gltf} rather than \texttt{click}. Each operation is also annotated with its execution procedure, which can later inform rollout guidance. We filter trivial or non-reusable candidates, then deduplicate, cluster, and aggregate operations across runs into a global \emph{operation pool} with supporting evidence.

\paragraph{Task instantiation.}
We then compose operations into diverse tasks grounded in concrete application states. Each task is conditioned on a real seed project, such as a \texttt{.blend} scene or \texttt{.odp} deck, together with its metadata, which defines the initial state $s_0$ and the objects available for manipulation. We sample and score candidate \emph{operation chains} and retain only compositions that are meaningful for the target seed. Difficulty is controlled by chain length and composition, ranging from single-operation edits to multi-step hybrid workflows. Each selected chain is compiled with its seed, supporting evidence, and mapped CLI tools into a \emph{task package} containing an instruction, initial state $s_0$, verifier, and guidance. The instruction is synthesized as a natural user goal rather than a sequence of low-level actions.

\paragraph{Agentic task refinement.}
Finally, we ground synthesized tasks in real execution before retaining them. A ReAct-style review agent~\cite{yao2022react} launches the application and performs a short multimodal interaction to check whether the instruction is feasible, unambiguous, and not already satisfied by the seed. When needed, it revises the instruction and guidance based on execution feedback. Valid tasks are retained, fixable tasks are revised, and invalid tasks are discarded,reducing hallucinated or infeasible task specifications before rollout.

\subsection{Path-Steer: Efficiency-Aware Hybrid Rollout}
\label{sec:rollout}
Path-Steer converts synthesized tasks into efficient, verified training trajectories through three stages: \emph{hybrid execution} enables GUI and CLI actions within a shared trajectory, \emph{efficient-path steering} guides agents toward appropriate modality choices, and \emph{trajectory harvesting} retains high-quality rollouts for post-training. Each task is attempted multiple times from fresh environment instances, with attempts executed in parallel for throughput.

\paragraph{Hybrid execution interface.}
We first provide agents with a unified interface for flexibly interleaving GUI and CLI actions. At each step, the agent emits either a GUI action or a CLI action. CLI actions are parsed against the application's tool registry, expanded into concrete commands with the current working-file path injected when needed, and executed in the VM. Their return code and truncated output are included in the next observation, allowing both modalities to operate seamlessly within one trajectory over the shared project state maintained by application adapters (\S\ref{sec:tool}).
\label{sec:exp}

\begin{figure*}[t]
    \centering
    \begin{subfigure}[b]{0.32\linewidth}
        \centering
        \includegraphics[width=\linewidth]{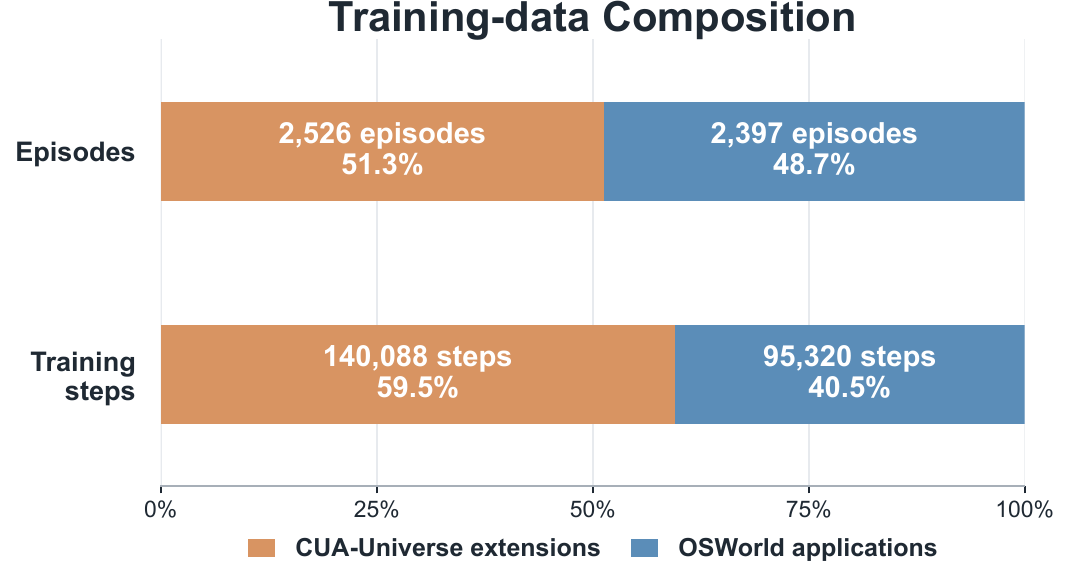}
        \caption{Overall distribution of episodes and training steps.}
        \label{fig:data-overall}
    \end{subfigure}
    \hfill
    \begin{subfigure}[b]{0.32\linewidth}
        \centering
        \includegraphics[width=\linewidth]{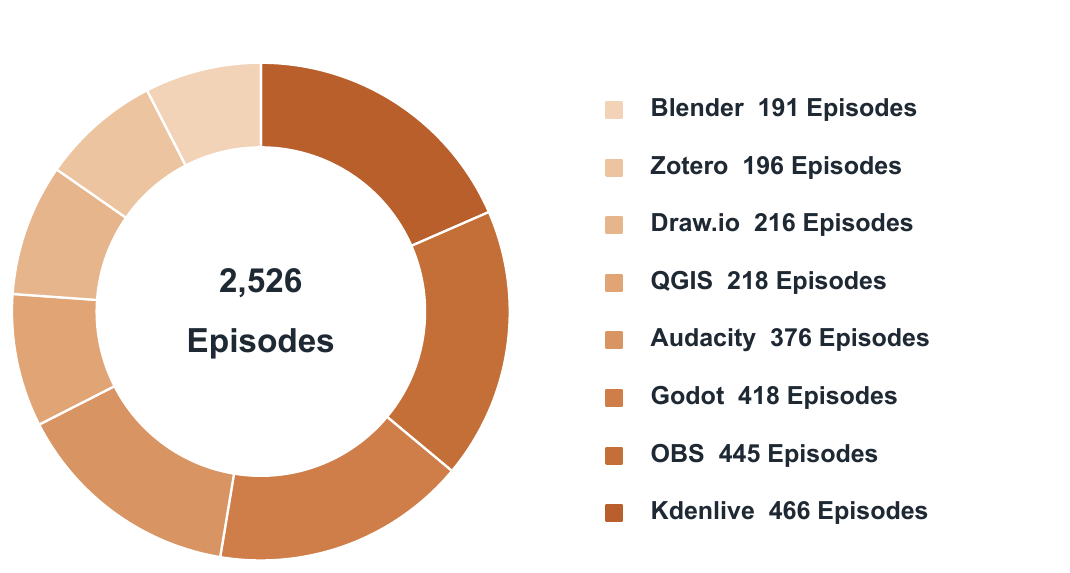}
        \caption{Task distribution across CUA-Universe extension applications.}
        \label{fig:data-cua}
    \end{subfigure}
    \hfill
    \begin{subfigure}[b]{0.32\linewidth}
        \centering
        \includegraphics[width=\linewidth]{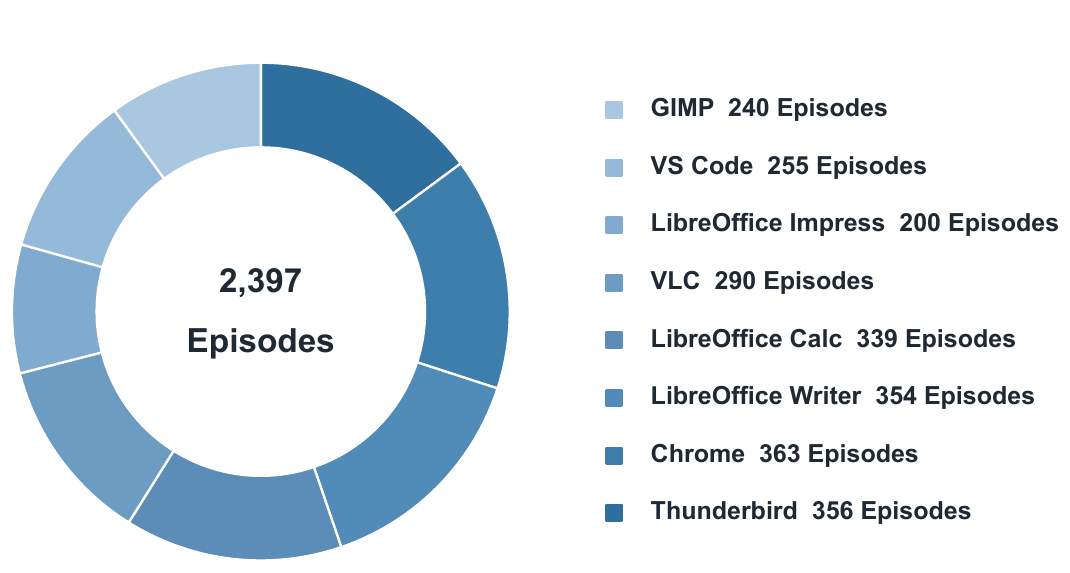}
       \caption{Task distribution across OSWorld applications.}
        \label{fig:data-osworld}
    \end{subfigure}
    \caption{\textbf{Training-data composition.} Our pipeline scales to diverse applications and produces substantial training data, with 4{,}923 episodes and approximately 235K training steps across all 16 applications. The balanced coverage of CUA-Universe extensions and OSWorld applications further demonstrates the diversity and scalability of the generated data.}
    \label{fig:data-composition}
\end{figure*}
% caption 加名词短语

\paragraph{Efficient-path steering.}
We then guide rollouts toward more efficient and appropriate modality choices. From each task's operation chain, we derive a \emph{hybrid execution prior} indicating when an operation is better suited to the CLI, such as batch, precise, or high-throughput operations, or to the GUI, such as operations depending on visual layout or interface state. The prior provides lightweight modality-level guidance without specifying low-level actions, reducing inefficient GUI interaction and brittle CLI scripting while yielding shorter and less redundant trajectories.

\paragraph{Scoring and trajectory harvesting.}
Finally, we score each completed rollout with the task verifier $V(\zeta)\in[0,1]$, implemented as a VLM judge over the trajectory. Verified rollouts are serialized into step-level and trajectory-level records containing observations, reasoning, GUI and CLI actions with their returns, screenshots, and final scores. We retain high-scoring hybrid trajectories as directly reusable supervision for subsequent post-training.

\section{Experiments}
\paragraph{Training setup.}
App installation and CLI-tool construction (App-Forge) are driven by a Codex coding agent (GPT-5.6), while operation abstraction and hybrid task synthesis (Task-Weave) are performed by Kimi~K2.5~\cite{team2026kimi}; task success throughout the pipeline is scored by a VLM judge (GPT-5.4~\cite{openai2026gpt54}). Our training data is generated by rolling out the same Kimi~K2.5 backbone under Path-Steer and keeping trajectories that pass the judge at a score threshold of $0.75$, yielding 4{,}923 verified episodes ($\sim$235K step-level records; Figure~\ref{fig:data-composition}). We then fine-tune Qwen3.5-9B~\cite{qwen3.5} with LoRA on these step-level trajectories for 3 epochs using the ms-swift~\cite{zhao2024swiftascalablelightweightinfrastructure} framework, on $8\times$ A100 GPUs in roughly two days; full training details are in Appendix~\ref{app:training}.

% ---------------------------------------------------------------------
% Main table  (two rows per model, single-line metric names)
% ---------------------------------------------------------------------
\begin{table*}[h]
\centering
\small
\setlength{\tabcolsep}{5pt}
\renewcommand{\arraystretch}{1.15}
\setlength{\aboverulesep}{0pt}
\setlength{\belowrulesep}{0pt}
\setlength{\extrarowheight}{0pt}

\begin{tabular}{ll
  >{\columncolor{CUAOrange}[\tabcolsep][\tabcolsep]}c
  >{\columncolor{CUAOrange}[\tabcolsep][\tabcolsep]}c
  >{\columncolor{CUAOrange}[\tabcolsep][\tabcolsep]}c
  >{\columncolor{OSWorldBlue}[\tabcolsep][\tabcolsep]}c
  >{\columncolor{OSWorldBlue}[\tabcolsep][\tabcolsep]}c
  >{\columncolor{OSWorldBlue}[\tabcolsep][\tabcolsep]}c
  >{\columncolor{OSWorldBlue}[\tabcolsep][\tabcolsep]}c
  >{\columncolor{OSWorldBlue}[\tabcolsep][\tabcolsep]}c}
\toprule
\multirow{2}{*}{\textbf{Model}} & \multirow{2}{*}{\textbf{Mode}}
 & \multicolumn{3}{>{\columncolor{CUAOrange}[\tabcolsep][\tabcolsep]}c}{\textbf{CUA-Verse}}
 & \multicolumn{5}{>{\columncolor{OSWorldBlue}[\tabcolsep][\tabcolsep]}c}{\textbf{OSWorld}} \\
\cmidrule(lr){3-5} \cmidrule(lr){6-10}
 &  & \textbf{Score} & \textbf{Steps} & \textbf{Token (K)}
    & \textbf{SR (\%)}
    & \textbf{Steps $\downarrow$}
    & \textbf{Step Gain $\uparrow$}
    & \textbf{Token(K) $\downarrow$}
    & \textbf{Token Gain $\uparrow$} \\
\midrule

                                & GUI
                                &                                  &                        &                       
                                & 53.7 & 27.2 &                        & 224.8 &                        \\
\multirow{-2}{*}{Kimi K2.5}     & GUI+CLI
                                & \multirow{-2}{*}{0.522}
                                & \multirow{-2}{*}{21.1}
                                & \multirow{-2}{*}{275}
                                & 54.5 & 28.9
                                & \multirow{-2}{*}{1.15}
                                & 277.3
                                & \multirow{-2}{*}{1.09} \\
\midrule

                                & GUI
                                &                                  &                        &                       
                                & 55.7 & 28.4 &                        & 188.0 &                        \\
\multirow{-2}{*}{Seed2.1 Pro}   & GUI+CLI
                                & \multirow{-2}{*}{0.599}
                                & \multirow{-2}{*}{40.1}
                                & \multirow{-2}{*}{449}
                                & 59.0 & 27.7
                                & \multirow{-2}{*}{1.10}
                                & 215.6
                                & \multirow{-2}{*}{0.94} \\
\midrule

                                & GUI
                                &                                  &                        &                       
                                & 66.8 & 14.7 &                        & 595.0 &                        \\
\multirow{-2}{*}{GPT-5.5}       & GUI+CLI
                                & \multirow{-2}{*}{0.768}
                                & \multirow{-2}{*}{23.1}
                                & \multirow{-2}{*}{193}
                                & 68.0 & 19.0
                                & \multirow{-2}{*}{1.55}
                                & 550.3
                                & \multirow{-2}{*}{1.58} \\
\midrule

                                & GUI
                                &                                  &                        &                       
                                & 21.7 & 58.6 &                        & 521.9 &                        \\
\multirow{-2}{*}{Qwen3.5-9B}    & GUI+CLI
                                & \multirow{-2}{*}{0.189}
                                & \multirow{-2}{*}{56.2}
                                & \multirow{-2}{*}{643}
                                & 24.6 & 54.1
                                & \multirow{-2}{*}{2.63}
                                & 596.1
                                & \multirow{-2}{*}{2.03} \\
\midrule

                                & GUI
                                &                                  &                        &                       
                                & 40.6 & 34.8 &                        & 325.6 &                        \\
\multirow{-2}{*}{EvoCUA-8B}     & GUI+CLI
                                & \multirow{-2}{*}{0.330}
                                & \multirow{-2}{*}{41.7}
                                & \multirow{-2}{*}{377}
                                & 43.3 & 42.1
                                & \multirow{-2}{*}{1.58}
                                & 408.0
                                & \multirow{-2}{*}{1.35} \\
\midrule

                                & GUI
                                &                                  &                        &                       
                                & 23.4 & 39.6 &                        & 325.7 &                        \\
\multirow{-2}{*}{\textbf{Ours}} & GUI+CLI
                                & \multirow{-2}{*}{0.582}
                                & \multirow{-2}{*}{35.2}
                                & \multirow{-2}{*}{255}
                                & 40.2 & 28.6
                                & \multirow{-2}{*}{2.35}
                                & 286.5
                                & \multirow{-2}{*}{1.79} \\
\bottomrule
\end{tabular}

\caption{\textbf{Combined performance on CUA-Verse and OSWorld.}
\emph{CUA-Verse}: 8 apps $\times$ 20 tasks; Score is the average VLM judge score, and Token is the average number of tokens per episode in thousands.
\emph{OSWorld}: 244-task controlled scope excluding \texttt{os} and multi-app tasks, comparing GUI and GUI+CLI execution. Token/task is reported in thousands. \textbf{Step Gain} and \textbf{Token Gain} are the mean per-task GUI-to-GUI+CLI cost ratios over jointly solved tasks. A gain of $k\times$ indicates that GUI+CLI uses $k$ times fewer steps or tokens on the same successfully solved tasks. }
\label{tab:merged}
\end{table*}

\subsection{Evaluation on CUA-Verse}
\label{sec:exp-cuaverse}

\begin{figure}[h]
    \centering
    \includegraphics[width=1\linewidth]{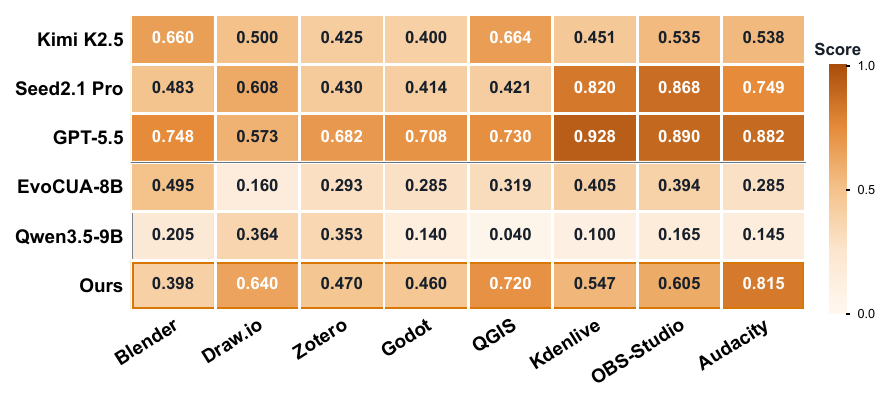}
    \caption{\textbf{Application-level performance on the eight CUA-Verse applications.} Each cell is the score for one model–application pair; darker shading is better. The horizontal rule separates proprietary (top) from open-source (bottom) models. Ours surpasses its Qwen3.5-9B  on all eight applications, and is strongest on audio/video apps (Audacity, OBS).}
    \label{fig:perapp}
\end{figure}

\paragraph{Setting.}
We evaluate on \textbf{CUA-Verse}, a held-out benchmark of 160 hybrid GUI+CLI tasks (eight professional desktop applications $\times$ 20 tasks) synthesized by the CUA-Universe pipeline. Its tasks are disjoint from all training data, though the eight applications themselves are in-domain; it measures an agent's ability to complete GUI+CLI hybrid tasks, exposing both a GUI and a CLI interface so the agent must coordinate the two over a shared application state to solve each task. The eight applications (Blender, Draw.io, Zotero, Godot, QGIS, Kdenlive, OBS, Audacity) are each scored by a unified VLM judge using GPT-5.4~\cite{openai2026gpt54}; we additionally report steps and per-episode token as efficiency metrics. We compare against three proprietary models (Kimi~K2.5, Seed2.1~Pro, GPT-5.5) and three open-source models (Qwen3.5-9B, EvoCUA-8B, and Ours). 
% Ours is initialized from Qwen3.5-9B and fine-tuned on high-scoring Kimi~K2.5 trajectories via rejection sampling. 模型的训练统统写前面
\paragraph{Results.}
Three observations stand out (Table~\ref{tab:merged}, Fig.~\ref{fig:perapp}). \textbf{(1) Distillation lifts a 9B model to closed-source level:} Ours reaches $0.582$ Score, the best among open-source models---surpassing Kimi~K2.5 ($0.522$) and trailing only the proprietary Seed2.1~Pro ($0.599$) and GPT-5.5 ($0.768$). Since it is trained only on the teacher's successful trajectories, it learns the upper tail and exceeds the teacher's mean on CUA-Verse, consistent with STaR~\cite{zelikman2022star} and ReST~\cite{gulcehre2023reinforced}. \textbf{(2) The gains are clean and efficient:} against the identical Qwen3.5-9B ($0.189$), Score improves $\sim$$3\times$ with $37\%$ fewer steps and $60\%$ fewer tokens ($255$K vs.\ $643$K per episode), yielding the best accuracy--cost trade-off among open models. % TODO(fig:cost): 精度-成本图尚未插入(候选 osworld_efficiency_frontier.pdf);插入后把上句改回 "...among open models (Fig.~\ref{fig:cost})." 并给该图打 \label{fig:cost}
\textbf{(3) Capability is structured:} Ours is strongest on audio/video apps (Audacity $0.815$, OBS $0.605$) and weaker on 3D/spatial ones (Blender $0.398$, Godot $0.460$); among open 8--9B models it dominates EvoCUA-8B ($0.330$ avg.) everywhere except Blender ($0.495$), indicating headroom in 3D data coverage.

% ---------------------------------------------------------------------
% \begin{table}[h]
% \centering
% \setlength{\tabcolsep}{8pt}\renewcommand{\arraystretch}{1.15}
% \begin{tabular}{lccc}
% \toprule
% \textbf{Model} & \textbf{SR} & \textbf{Steps} & \textbf{Token(K)} \\
% \midrule
% % \multicolumn{4}{l}{\emph{Closed-source}} \\
% Kimi 2.5          & 0.522                     & 21.1 & 275 \\
% Seed2.1 Pro  & 0.599                     & 40.1 & 449 \\
% GPT-5.5           & \textbf{0.768}            & 23.1 & 193 \\
% \midrule
% % \multicolumn{4}{l}{\emph{Open-source}} \\
% Qwen3.5-9B  & 0.189                     & 56.2 & 643 \\
% EvoCUA-8B         & 0.330            & 41.7 & 377 \\
% \rowcolor{OursRow}\textbf{Ours} & \textbf{0.582} & 35.2 & 255 \\
% \bottomrule
% \end{tabular}
% \caption{Overall performance and efficiency on CUA-Verse (8 apps $\times$ 20 tasks). SR is the average VLM judge score; Token is the average token per episode. }
% \label{tab:overall}
% \end{table}

\subsection{Transfer to OSWorld}
\label{sec:exp-osworld}

\begin{figure*}[t]
\centering
% ============ 左：OSWorld 图 ============
\begin{minipage}[t]{0.66\linewidth}
    \vspace{0pt}          % 强制顶边对齐
    \centering
    \includegraphics[width=\linewidth]{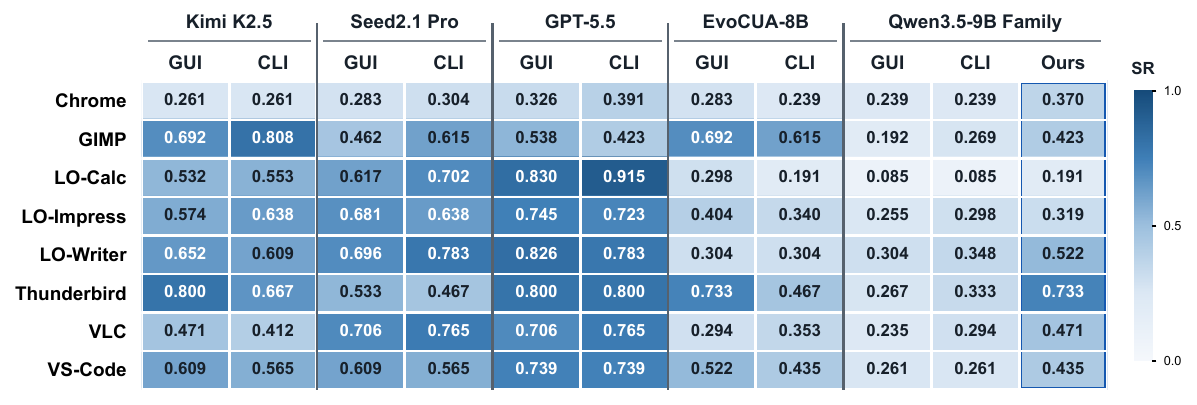}
    \captionof{figure}{\textbf{Per-application OSWorld scores across models and interfaces (GUI vs.\ GUI+CLI).} Cells are heat-shaded by score. Adding the CLI yields Ours (rightmost) its largest gains, converting +41 previously-failed GUI tasks into successes.}
    \label{fig:frontier}
\end{minipage}
\hfill
% ============ 右：OSWorld-MCP 表 ============
\begin{minipage}[t]{0.32\linewidth}
    \vspace{7pt}          % 强制顶边对齐
    \centering
    \small
    \setlength{\tabcolsep}{3.2pt}
    \renewcommand{\arraystretch}{1.15}
    \begin{tabular}{lrrrr}
    \toprule
    \textbf{Model} & \multicolumn{1}{c}{\textbf{SR}$\uparrow$} & \multicolumn{1}{c}{\textbf{TIR}$\uparrow$} & \multicolumn{1}{c}{\textbf{ACS}$\downarrow$} & \multicolumn{1}{c}{\textbf{Tok}$\downarrow$} \\
    \midrule
    Kimi K2.5$^{\dagger}$      & 35.25 & 30.33 & 27.52 & 85.21 \\
    Seed2.1 Pro$^{\dagger}$    & 26.23 & 22.54 & 40.62 & 99.30 \\
    GPT-5.5$^{\dagger}$        & 29.92 & 25.00 & 27.06 & 54.62 \\
    \midrule
    Qwen3.5-9B                 & 20.90 & 10.66 & 37.22 & 125.68 \\
    \rowcolor{OursRow}\textbf{Ours} & 28.69 & 23.36 & 27.25 & 87.95 \\
    \bottomrule
    \end{tabular}
    \captionof{table}{\textbf{Generalization on OSWorld-MCP} (244-task subset, exclude \ \texttt{os}/\texttt{multi\_apps}). $\uparrow$ higher / $\downarrow$ lower is better; Tokens in M. $^{\dagger}$: unified prompt, reference value.}
    \label{tab:mainmcp}
\end{minipage}
\end{figure*}

\paragraph{Setting.}

We evaluate whether training on CUA-Universe improves general computer-use capability even under GUI-only execution, and whether providing CLI access yields further gains through learned hybrid orchestration. We evaluate on OSWorld under a controlled \textbf{244-task} protocol that excludes the \texttt{os} and multi-app splits and scores every task with the official verifier; \emph{SR} is the fraction of tasks with reward $1$. We compare two action interfaces under \emph{identical} task text, environment setup, and verifier: in the \textbf{GUI} setting the agent acts purely through screen coordinates, and in the \textbf{GUI+CLI} setting it additionally receives a native \texttt{execute\_cli} tool with per-application command definitions and a single injected sentence giving the task's input-file path. All agents share a 60-step budget; observation history is each baseline's default, while our Qwen-family models use a 3-frame history with at most 4 images per step. \emph{Steps} is the mean number of model decision calls per trajectory and \emph{Token/task} the mean input$+$output tokens per task (thousands). Full harness and hardware details are in Appendix~\ref{app:compute}.

\paragraph{Interface metrics.}
Because GUI and GUI+CLI are evaluated on the same 244 tasks, we report paired measures of both success and efficiency gains from adding CLI access. \emph{Success gain} is the net number of tasks newly solved by GUI+CLI, computed as $(\mathrm{SR}_{\text{GUI+CLI}}-\mathrm{SR}_{\text{GUI}})\times244$, and is reported directly from the SR results. \textbf{Step Gain} and \textbf{Token Gain} are computed only on tasks solved by both interfaces, using the mean per-task ratio of GUI to GUI+CLI cost for steps and tokens, respectively. A value of $k\times$ means that GUI+CLI reaches the same successful outcome with $k$ times fewer steps or tokens, while averaging per-task ratios prevents a few long trajectories from dominating the metric.

\begin{table*}[!t]
\centering
\begin{tabular}{lccccc}
\toprule
Group & Accept Rate ($\geq 0.75$) & Mean score & Avg steps & Avg tokens & Avg cost \\
\midrule
Kimi K2.5 \emph{w/ Path-Steer}     & 0.51 & 0.71 & 22.75 & 331{,}988 & \$0.26 \\
Kimi K2.5 \emph{w/o Path-Steer}    & 0.44 & 0.63 & 26.68 & 385{,}107 & \$0.31 \\
Seed2.1 Pro \emph{w/ Path-Steer}  & 0.54 & 0.75 & 25.53 & 263{,}500 & \$0.29 \\
Seed2.1 Pro \emph{w/o Path-Steer} & 0.45 & 0.67 & 28.78 & 303{,}653 & \$0.33 \\
\bottomrule
\end{tabular}
\caption{\textbf{Rollout efficiency of Path-Steer (enabled vs.\ disabled)}, for the Kimi K2.5 data-generation backbone and Seed2.1 Pro (a cross-backbone check). Accept Rate is the fraction of the 320 tasks scoring $\geq 0.75$; tokens are averaged per task.}

\label{tab:rollout}
\end{table*}

\paragraph{Results and analysis.}
Table~\ref{tab:merged} and Figure~\ref{fig:frontier} summarize the comparison, from which we draw five findings. \textbf{(1) Task-specific distillation lifts a 9B open model to near open-SOTA at the lowest cost:} with GUI+CLI our model reaches \textbf{40.2\%} SR---up \textbf{+16.8} points from its GUI-only 23.4\% and within 3 points of EvoCUA-8B (43.3\%)---while spending the fewest tokens per task (286.5k) and the fewest steps (28.6) of any agent, and more than doubling the untuned Qwen3.5-9B (24.6\%) under the same interface. \textbf{(2) The CLI interface is where our model's gain concentrates---by far the largest net solved-task gain of any agent:} adding CLI converts \textbf{+41} previously-failed GUI tasks into successes ($23.4\rightarrow40.2\%$), against only $+3$ to $+8$ for every other model, none of which gains more than eight tasks. \textbf{(3) For strong closed backbones the interface alone helps only marginally:} GPT-5.5 ($+3$), Seed2.1 ($+8$) and EvoCUA-8B ($+7$) improve modestly and Kimi~K2.5 is essentially flat ($+2$)---exposing a CLI the model was not trained to exploit yields little without task-specific tuning. \textbf{(4) When CLI helps, it is also cheaper:} on jointly-solved tasks the CLI interface uses fewer steps for every model ($1.10\text{--}2.63\times$) and fewer tokens for most (up to $2.03\times$); our model reaches the same successes with $2.35\times$ fewer steps. The lone exception is Seed2.1, whose sub-$1\times$ token efficiency shows the CLI interface can trade extra tokens for its accuracy gain. \textbf{(5) Overall ranking:} among closed agents GPT-5.5 (68.0\%) $>$ Seed2.1 (59.0\%) $>$ Kimi~K2.5 (54.5\%); among open models EvoCUA-8B (43.3\%) leads with our model (40.2\%) a close second, both far above the base.

\subsection{Generalization to OSWorld-MCP}

\paragraph{Setting.}
We evaluate whether CUA-Universe teaches transferable cross-modality orchestration, rather than benchmark-specific interaction patterns, on the held-out OSWorld-MCP benchmark~\citep{jia2025osworld}, which augments OSWorld with 158 MCP tools and allows agents to freely combine GUI actions and tool calls. We report \emph{Score} (task accuracy), \emph{Strict SR} (perfect-score rate), \emph{TIR} (accuracy of tool-use decisions), and \emph{ACS} (average completion steps, lower is better), together with tokens per run as an additional efficiency metric. Evaluation uses 244 tasks after excluding \texttt{os} and \texttt{multi\_apps}, including 159 tool-beneficial and 85 non-tool-beneficial tasks, with $\mathrm{max\_steps}=50$ and $\mathrm{history\_n}=3$. We compare against Kimi~K2.5, Seed2.1~Pro, GPT-5.5, and Qwen3.5-9B.

% \begin{table}[t]
% \centering
% \small
% \setlength{\tabcolsep}{4.9pt}
% \renewcommand{\arraystretch}{1.15}
% \begin{tabular}{lrrrrr}
% \toprule
% \textbf{Model} & \multicolumn{1}{c}{\textbf{Score}$\uparrow$} & \multicolumn{1}{c}{\textbf{Strict SR}$\uparrow$} & \multicolumn{1}{c}{\textbf{TIR}$\uparrow$} & \multicolumn{1}{c}{\textbf{ACS}$\downarrow$} & \multicolumn{1}{c}{\textbf{Tokens}$\downarrow$} \\
% \midrule
% Kimi K2.5$^{\dagger}$ & \textbf{36.05} & \textbf{35.25} & \textbf{30.33} & 27.52 & 85.21 \\
% Seed2.1 Pro$^{\dagger}$ & 27.03 & 26.23 & 22.54 & 40.62 & 99.30 \\
% GPT-5.5$^{\dagger}$ & 30.73 & 29.92 & 25.00 & \textbf{27.06} & \textbf{54.62} \\
% \midrule
% Qwen3.5-9B  & 21.67 & 20.90 & 10.66 & 37.22 & 125.68 \\
% \rowcolor{OursRow}\textbf{Ours} & 29.51 & 28.69 & 23.36 & 27.25 & 87.95 \\
% \bottomrule
% \end{tabular}
% \caption{\textbf{Generalization performance on OSWorld-MCP} (244-task subset, excluding \texttt{os} and \texttt{multi\_apps}). Higher is better for Score, SR, and TIR; lower is better for ACS and tokens. $^{\dagger}$: the official repository provides no Seed-specific prompt, so we use the unified prompt and report it as a reference value.}
% \label{tab:mainmcp}
% \end{table}

\paragraph{Results.}
Three observations stand out (Table~\ref{tab:mainmcp}). \textbf{(1) Hybrid interaction skills transfer to an unseen MCP interface:} our model is trained only with application-specific CLI tools in CUA-Universe and never sees the MCP action space, yet it improves SR from $20.90\%$ to $28.69\%$ ($+7.79$ points) and more than doubles TIR from $10.66\%$ to $23.36\%$ over the base, showing that the learned tool-use behavior transfers beyond the training format. \textbf{(2) The transferred capability is competitive with substantially larger models:} a single 9B model surpasses Seed2.1~Pro on Score ($29.51\%$ vs.\ $27.03\%$) and approaches GPT-5.5 and Kimi~K2.5, narrowing the gap to far larger closed-source models. \textbf{(3) Generalization remains efficient:} Ours reaches an ACS of $27.25$ with $87.95$M tokens, reducing steps by $27\%$ and tokens by $30\%$ relative to its base. These results show that the orchestration capability learned in CUA-Universe transfers to MCP without sacrificing efficiency.

\subsection{Rollout Efficiency}
\label{sec:exp-rollout}

\paragraph{Setting.}
We evaluate Path-Steer during data generation to measure whether explicit modality guidance improves both rollout quality and efficiency. On a fixed set of 320 synthesized tasks, we compare rollouts with and without Path-Steer while keeping the hybrid GUI+CLI interface unchanged. We use Kimi~K2.5, our data-generation backbone, and additionally evaluate Seed2.1~Pro to test whether the effect generalizes across models. We report \emph{Accept Rate}, the fraction of trajectories scoring at least $0.75$, and \emph{Mean Score} for trajectory quality, together with average \emph{steps}, \emph{tokens}, and estimated per-task \emph{cost} for efficiency. Since both settings have identical CLI access, this comparison isolates the contribution of steering itself.

\paragraph{Results.}
Three observations stand out (Table~\ref{tab:rollout}). \textbf{(1) Path-Steer improves trajectory quality and efficiency simultaneously:} on Kimi~K2.5, it raises Accept Rate from $0.44$ to $0.51$ and Mean Score from $0.63$ to $0.71$, while reducing steps from $26.68$ to $22.75$, tokens from $385$K to $332$K, and per-task cost from $\$0.31$ to $\$0.26$. This indicates that explicit modality guidance produces shorter and higher-quality hybrid trajectories by reducing inefficient GUI interaction and brittle CLI scripting. \textbf{(2) The effect generalizes across backbones:} Seed2.1~Pro shows the same pattern, with Accept Rate improving from $0.45$ to $0.54$ and Mean Score from $0.67$ to $0.75$, together with lower steps, tokens, and cost. \textbf{(3) Steering, rather than CLI access alone, drives the gains:} the \emph{w/o Path-Steer} baseline already has access to the same CLI tools, so the consistent improvements isolate the contribution of modality steering itself. Step-by-step comparisons between the two settings are provided in Appendix~\ref{app:steer-path}.

\section{Conclusion}
\label{sec:conclusion}
We presented CUA-Universe as a step toward a different way of scaling computer-use agents: scaling the environments and interaction spaces from which agents learn, rather than relying only on larger models or more GUI-only trajectories. By turning real desktop software into shared-state GUI+CLI environments, CUA-Universe provides a scalable source of hybrid tasks and trajectories that teach agents not only how to act, but how to orchestrate complementary interfaces efficiently. The resulting 9B model shows that this capability is learnable and transferable, with strong gains on CUA-Verse and OSWorld and further generalization to the unseen tool interface of OSWorld-MCP. More broadly, our results suggest that hybrid environment construction can become a new axis for training computer-use agents, where each additional application expands the space of tasks, tools, and interaction strategies available for learning. We hope this shifts CUA development from collecting increasingly large amounts of single-modality behavior toward building scalable environments that continuously generate richer supervision for more capable, efficient, and general computer-use agents.

% \section{Limitations}
% \label{sec:limitations}
% CUA-Universe currently synthesizes single-application tasks; extending it to cross-application workflows over multi-app workspaces---where state is carried across several applications---is left to future work. We use the harvested trajectories for supervised fine-tuning only, so the learned policy is bounded by its data-generation backbone; leveraging the pipeline's verifiers as rewards for reinforcement learning is a natural next step. Task success is scored by a VLM judge rather than per-task programmatic checks, which may introduce label noise despite our score threshold. Finally, application adaptation assumes software that is open-source or scriptable enough to expose a command-line surface, and our environments target desktop Linux; broadening to closed-source or non-desktop platforms remains open.

% Required packages:
% \usepackage{booktabs}
% \usepackage{array}
% \usepackage[table]{xcolor}

\clearpage
\bibliography{aaai2026}

\clearpage
\appendix

\section{Additional Experiments}

\subsection{Action-Modality Behavior on CUA-Verse}
\label{app:action-modality}
\paragraph{Setting.}
Beyond overall performance, we examine how different models use the two modalities on CUA-Verse. CUA-Verse is explicitly constructed from hybrid tasks whose reference solutions span both GUI and CLI operations over a shared application state, including visually grounded interactions that cannot be reduced to command execution alone. From the stored trajectories, we classify each nonterminal action as CLI or GUI and report \emph{CLI\,\%}, the fraction of executed GUI$+$CLI actions that use the CLI. We report it alongside \emph{Steps} and the CUA-Verse \emph{Score} to characterize the relationship among modality choice, efficiency, and performance.

\begin{table}[h]
\centering
\small
\setlength{\tabcolsep}{6pt}
\renewcommand{\arraystretch}{1.12}

\begin{tabular}{lccr}
\toprule
\textbf{Model} & \textbf{Steps} & \textbf{CLI\,\%} & \textbf{Score} \\
\midrule
Kimi~K2.5       & 21.1 & \phantom{0}30.3 & 0.522 \\
Seed2.1 Pro     & 40.1 & \phantom{0}49.2 & 0.599 \\
GPT-5.5         & 23.1 & 100.0 & 0.768 \\
\midrule
Qwen3.5-9B      & 56.2 & \phantom{00}0.0 & 0.189 \\
EvoCUA-8B       & 41.7 & \phantom{00}3.0 & 0.330 \\
\rowcolor{OursRow}
Ours             & 35.2 & \phantom{0}25.3 & 0.582 \\
\bottomrule
\end{tabular}

\caption{\textbf{Action-modality behavior on CUA-Verse}, averaged over the eight applications. \emph{Steps} is the mean executed action count, \emph{CLI\,\%} is the CLI share of executed GUI$+$CLI actions, and \emph{Score} is the average VLM-judge score. Greater CLI use often accompanies higher efficiency and performance, but CLI usage alone is insufficient: even GPT-5.5, which executes entirely through CLI in these trajectories, does not fully solve the benchmark, reflecting the visually grounded demands of CUA-Verse.}
\label{tab:action-modality}
\end{table}

\paragraph{Results.}
Table~\ref{tab:action-modality} shows that effective CLI use is important, but the key capability is cross-modality orchestration rather than maximizing CLI usage. \textbf{(1) The untuned base remains GUI-bound.} Qwen3.5-9B issues no CLI actions, uses nearly the full 60-step budget, and obtains the lowest Score ($0.189$), reflecting inefficient GUI-only execution. \textbf{(2) Hybrid training changes interaction behavior.} With the same backbone, our model raises CLI usage to $25.3\%$, reduces mean steps by $37\%$, and increases Score from $0.189$ to $0.582$, showing that CUA-Universe teaches the model to exploit CLI operations while retaining GUI interaction when needed. \textbf{(3) More CLI is not itself sufficient.} GPT-5.5 uses CLI for all recorded actions and achieves the highest Score ($0.768$), yet still falls well short of perfect performance. Conversely, EvoCUA-8B remains almost entirely GUI-based despite having access to the CLI and obtains only $0.330$. Together, these results suggest that CUA-Verse rewards the ability to select and coordinate modalities according to the task, rather than simply favoring either GUI or CLI in isolation.

\subsection{Out-of-Domain Training Transfer}
\label{app:ood-transfer}

\paragraph{Setting.} To test whether the LoRA gains reflect a transferable CLI-usage capability rather than memorization of the evaluation applications, we train an \textbf{8-app OOD-only LoRA} exclusively on applications that are disjoint from the eight OSWorld evaluation applications and evaluate it zero-shot under the same controlled 244-task GUI+CLI scope. We compare it against the untuned Qwen3.5-9B and our full \textbf{16-app LoRA}, whose training mixture additionally includes the eight OSWorld application domains.

\begin{table}[t]
\centering
\small
\setlength{\tabcolsep}{6pt}
\renewcommand{\arraystretch}{1.12}
\begin{tabular}{lccc}
\toprule
\textbf{App} & \textbf{Base} & \makecell{\textbf{8-app LoRA}\\\textbf{(OOD-only)}} & \makecell{\textbf{16-app LoRA}\\\textbf{(+in-domain)}} \\
\midrule
Chrome      & 23.9 & 26.1 & 37.0 \\
GIMP        & 26.9 & 30.8 & 42.3 \\
Calc        & 8.5  & 10.6  & 18.8 \\
Impress     & 27.7 & 25.8 & 31.8 \\
Writer      & 34.8 & 39.1 & 52.2 \\
Thunderbird & 33.3 & 46.7 & 73.3 \\
VLC         & 29.4 & 31.4 & 37.4 \\
VS\,Code    & 26.1 & 34.8 & 43.5 \\
\midrule
\textbf{All apps} & 24.2 & $27.4\,(+3.2)$ & $\mathbf{40.2\,(+16.0)}$ \\
\bottomrule
\end{tabular}
\caption{\textbf{Out-of-domain training transfer on OSWorld} under the controlled 244-task GUI+CLI scope. Values are success rates in percent. The 8-app OOD-only LoRA is trained on cua-universe extensions applications disjoint from the OSWorld applications, whereas the full 16-app LoRA is trained on both. Parenthesized values in the final row denote absolute percentage-point gains over the untuned Qwen3.5-9B. }
\label{tab:osworld-ood}
\end{table}

\paragraph{Results.} Three observations stand out (Table~\ref{tab:osworld-ood}). \textbf{(1) Hybrid CLI-usage skills transfer across applications:} despite being trained exclusively on applications disjoint from the OSWorld evaluation applications, the 8-app OOD-only LoRA improves over the untuned base on seven of the eight applications and raises the overall success rate from $24.2\%$ to $27.4\%$ ($+3.2$ points). This zero-shot improvement indicates that the LoRA learns a partially application-agnostic capability for selecting and coordinating GUI and CLI actions rather than only memorizing application-specific commands. \textbf{(2) The transfer is heterogeneous across applications:} the largest out-of-domain gains occur on Thunderbird ($+13.4$ points) and VS\,Code ($+8.7$), followed by Writer ($+4.3$) and GIMP ($+3.9$); Impress is the only application on which performance decreases, by $1.9$ points. \textbf{(3) In-domain coverage substantially compounds the transferable gain:} the full 16-app LoRA improves over the base on all eight OSWorld applications and raises the overall success rate to $40.2\%$ ($+13.2$ points), showing that transferable hybrid-interaction skills and application-specific training data provide complementary benefits.

\section{Adapted Applications}
\label{app:apps}

CUA-Universe currently provides adapters for the 16 desktop applications listed in Table~\ref{tab:apps}, covering diverse software domains.
Eight applications are inherited from OSWorld, and eight additional applications are introduced by CUA-Universe.
Each application implements the uniform adapter interface described in \S\ref{sec:env}, exposing graphical, command-line, and programmatic interfaces that operate on the same underlying state.

\begin{table}[ht]
\centering
% \vspace{0.4em}
\scriptsize
\setlength{\tabcolsep}{3.5pt}
\renewcommand{\arraystretch}{1.08}
\begin{tabular}{>{\raggedright\arraybackslash}p{0.34\columnwidth}>{\raggedright\arraybackslash}p{0.37\columnwidth}>{\raggedright\arraybackslash}p{0.19\columnwidth}}
\toprule
\textbf{Application} & \textbf{Domain} & \textbf{Seed} \\
\midrule
\rowcolor{OSWorldBlue}\textcolor{OSWorldBlueText}{\textbf{OSWorld applications}} & & \\
\rowcolor{OSWorldBlue}Chrome              & Web browsing         & Profile             \\
\rowcolor{OSWorldBlue}GIMP                & Image editing        & \texttt{.xcf}       \\
\rowcolor{OSWorldBlue}LibreOffice Calc    & Spreadsheets         & \texttt{.ods}       \\
\rowcolor{OSWorldBlue}LibreOffice Impress & Presentations        & \texttt{.odp}       \\
\rowcolor{OSWorldBlue}LibreOffice Writer  & Word processing      & \texttt{.odt}       \\
\rowcolor{OSWorldBlue}Thunderbird         & Email                & Profile             \\
\rowcolor{OSWorldBlue}VLC                 & Media playback       & Media file          \\
\rowcolor{OSWorldBlue}VS Code             & Code editing         & Workspace           \\
% \addlinespace[0.3em]
\rowcolor{CUAOrange}\textcolor{CUAOrangeText}{\textbf{CUA-Universe extensions}} & & \\
\rowcolor{CUAOrange}Audacity   & Audio editing        & \texttt{.aup3}     \\
\rowcolor{CUAOrange}Blender    & 3D modelling         & \texttt{.blend}    \\
\rowcolor{CUAOrange}Draw.io    & Diagramming          & \texttt{.drawio}   \\
\rowcolor{CUAOrange}Godot      & Game development     & \texttt{.tscn}     \\
\rowcolor{CUAOrange}Kdenlive   & Video editing        & \texttt{.kdenlive} \\
\rowcolor{CUAOrange}OBS Studio & Screen recording     & \texttt{.json}     \\
\rowcolor{CUAOrange}QGIS       & Geospatial analysis  & \texttt{.qgz}      \\
\rowcolor{CUAOrange}Zotero     & Reference management & Library            \\
\bottomrule
\end{tabular}
\caption{\textbf{Desktop applications supported by CUA-Universe.} Blue rows indicate applications inherited from OSWorld, whereas orange rows indicate applications added by CUA-Universe.}
\label{tab:apps}
\end{table}

\section{Tool Construction}
\label{app:tools}

The tool layer converts each application's native automation surface---a scripting API, a project-file serializer, or an installed command-line binary---into a small, bounded vocabulary of subcommands that return JSON. A registry stores only metadata (tool ID, description, and the complete command template); the rollout runner executes the named command inside the benchmark VM and returns its exit code, stdout, and stderr to the next decision step. Where an application already ships a usable command-line interface, the registry wraps it directly; where it does not, a bounded harness operates on the native project artifact---or on the live application---so that every effect stays auditable in the same file the GUI opens.

\begin{figure}[t]
  \centering
  \includegraphics[width=\linewidth]{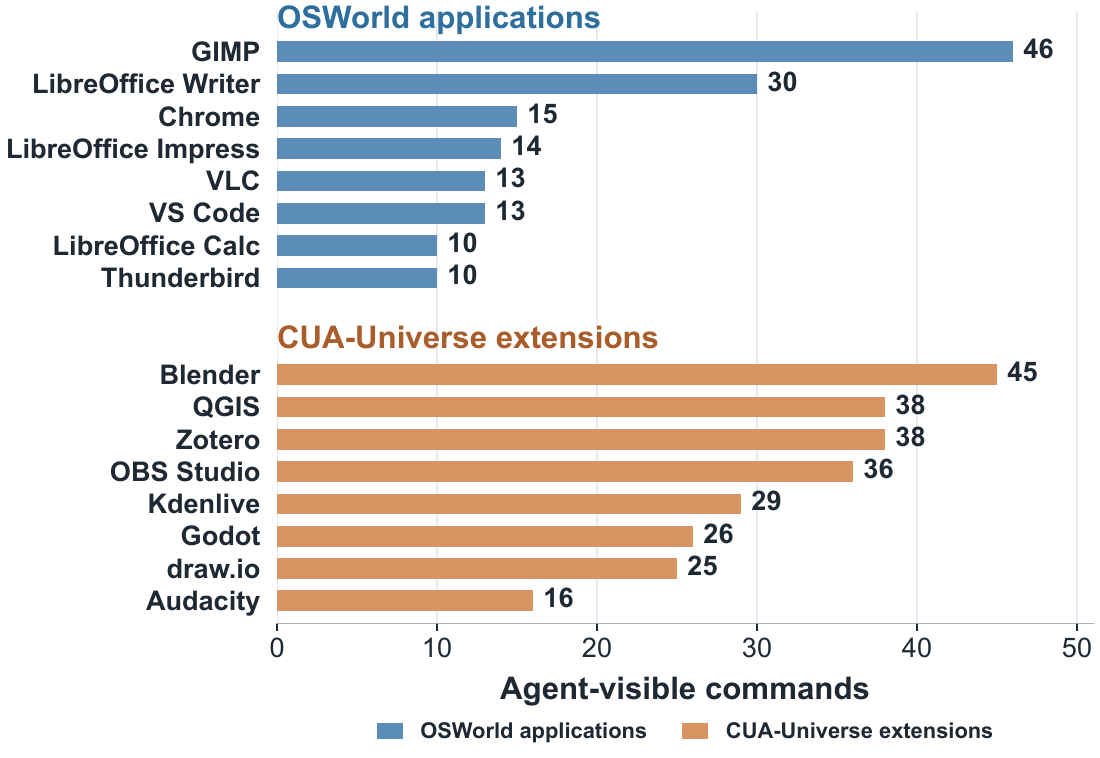}
  \caption{Agent-visible command count per application, grouped by source: the eight applications reused from OSWorld and the eight CUA-Universe extensions ($\sim$404 commands in total).}
  \label{fig:tool-coverage}
\end{figure}

\noindent The deployed snapshot spans \textbf{16 applications and $\sim$404 agent-visible commands}: 151 commands across the eight applications reused from OSWorld and 253 across the eight CUA-Universe extensions (Figure~\ref{fig:tool-coverage}). Command counts are not padded to a target range---they track each application's real capability surface. For brevity, the per-application sections below list tool \emph{IDs} grouped by command family.

\subsection{GIMP\,---\,46 commands}
\noindent The GIMP uses a live backend that drives the actual GUI application, while exposes the command interface and return results in JSON format.

\bigskip

\begin{toollist}
\item[\famlabel{project}] \tool{project\_new} \tool{project\_open} \tool{project\_save} \tool{project\_info} \tool{project\_json} \tool{project\_profiles}
\item[\famlabel{canvas}] \tool{canvas\_resize} \tool{canvas\_scale} \tool{canvas\_crop} \tool{canvas\_mode} \tool{canvas\_dpi} \tool{canvas\_info}
\item[\famlabel{layer}] \tool{layer\_new} \tool{layer\_add\_from\_file} \tool{layer\_list} \tool{layer\_remove} \tool{layer\_duplicate} \tool{layer\_move} \tool{layer\_set} \tool{layer\_flatten} \tool{layer\_merge\_down}
\item[\famlabel{filter}] \tool{filter\_add} \tool{filter\_list} \tool{filter\_list\_available} \tool{filter\_info} \tool{filter\_set} \tool{filter\_remove}
\item[\famlabel{draw}] \tool{draw\_text} \tool{draw\_rect}
\item[\famlabel{media}] \tool{media\_probe} \tool{media\_histogram} \tool{media\_list} \tool{media\_check}
\item[\famlabel{export}] \tool{export\_render} \tool{export\_presets} \tool{export\_preset\_info}
\item[\famlabel{live}] \tool{live\_status} \tool{live\_canvas\_info} \tool{live\_layer\_list} \tool{live\_draw\_text} \tool{live\_draw\_rect} \tool{live\_export\_render}
\item[\famlabel{session}] \tool{session\_undo} \tool{session\_redo} \tool{session\_history} \tool{session\_status}
\end{toollist}

\subsection{Blender\,---\,45 commands}
\noindent \sloppy A bounded bpy program runs in a single headless invocation, during which it has full read/write access to the real .blend artifact, allowing it to both inspect and alter the scene contents permanently.

\bigskip
\begin{toollist}
\item[\famlabel{scene}] \tool{scene\_new} \tool{scene\_open} \tool{scene\_save} \tool{scene\_info} \tool{scene\_profiles} \tool{scene\_json}
\item[\famlabel{object}] \tool{object\_add} \tool{object\_remove} \tool{object\_duplicate} \tool{object\_transform} \tool{object\_set} \tool{object\_list} \tool{object\_get}
\item[\famlabel{material}] \tool{material\_create} \tool{material\_assign} \tool{material\_set} \tool{material\_list} \tool{material\_get}
\item[\famlabel{modifier}] \tool{modifier\_list\_available} \tool{modifier\_info} \tool{modifier\_add} \tool{modifier\_remove} \tool{modifier\_set} \tool{modifier\_list}
\item[\famlabel{camera}] \tool{camera\_add} \tool{camera\_set} \tool{camera\_set\_active} \tool{camera\_list}
\item[\famlabel{light}] \tool{light\_add} \tool{light\_set} \tool{light\_list}
\item[\famlabel{animation}] \tool{animation\_keyframe} \tool{animation\_remove\_keyframe} \tool{animation\_frame\_range} \tool{animation\_fps} \tool{animation\_list\_keyframes}
\item[\famlabel{render}] \tool{render\_settings} \tool{render\_info} \tool{render\_presets} \tool{render\_execute} \tool{render\_script}
\item[\famlabel{session}] \tool{session\_status} \tool{session\_undo} \tool{session\_redo} \tool{session\_history}
\end{toollist}

\subsection{QGIS\,---\,38 commands}
\noindent \sloppy Bounded PyQGIS operations against a native \texttt{.qgs} project, with optional GUI sync, giving the agent full read and write access to layers, features, layouts, and processing pipelines.

\bigskip
\begin{toollist}
\item[\famlabel{project}] \tool{project\_new} \tool{project\_open} \tool{project\_info} \tool{project\_set\_crs} \tool{project\_save}
\item[\famlabel{layer}] \tool{layer\_create\_vector} \tool{layer\_list} \tool{layer\_info} \tool{layer\_remove} \tool{layer\_style\_simple} \tool{layer\_style\_graduated} \tool{layer\_style\_categorized} \tool{layer\_label\_simple} \tool{layer\_label\_buffer} \tool{layer\_category\_visibility} \tool{layer\_filter\_attribute}
\item[\famlabel{feature}] \tool{feature\_add} \tool{feature\_add\_point} \tool{feature\_list}
\item[\famlabel{layout}] \tool{layout\_create} \tool{layout\_list} \tool{layout\_info} \tool{layout\_remove} \tool{layout\_add\_map} \tool{layout\_add\_legend} \tool{layout\_legend\_remove\_layer} \tool{layout\_add\_label} \tool{layout\_sync\_extent}
\item[\famlabel{process}] \tool{process\_list} \tool{process\_help} \tool{process\_run}
\item[\famlabel{export}] \tool{export\_presets} \tool{export\_pdf} \tool{export\_image}
\item[\famlabel{session}] \tool{session\_status} \tool{session\_history}
\item[\famlabel{app / repl}] \tool{app\_open} \tool{repl}
\end{toollist}

\subsection{Zotero\,---\,38 commands}
\noindent \sloppy The seeded Zotero profile is accessed via its local database, browser connector, and Local API, enabling full management of collections, items, notes, and saved searches.

\bigskip
\begin{toollist}
\item[\famlabel{app}] \tool{app\_status} \tool{app\_version} \tool{app\_launch} \tool{app\_enable\_local\_api} \tool{app\_ping}
\item[\famlabel{collection}] \tool{collection\_list} \tool{collection\_find} \tool{collection\_tree} \tool{collection\_get} \tool{collection\_items} \tool{collection\_use\_selected} \tool{collection\_create} \tool{collection\_export}
\item[\famlabel{item}] \tool{item\_list} \tool{item\_find} \tool{item\_get} \tool{item\_children} \tool{item\_notes} \tool{item\_attachments} \tool{item\_file} \tool{item\_export} \tool{item\_citation} \tool{item\_bibliography} \tool{item\_context} \tool{item\_analyze} \tool{item\_add\_to\_collection} \tool{item\_move\_to\_collection}
\item[\famlabel{note}] \tool{note\_get} \tool{note\_add}
\item[\famlabel{search}] \tool{search\_list} \tool{search\_get} \tool{search\_items}
\item[\famlabel{tag / style}] \tool{tag\_list} \tool{tag\_items} \tool{style\_list}
\item[\famlabel{import / session}] \tool{import\_file} \tool{import\_json} \tool{session}
\end{toollist}

\subsection{OBS\,---\,36 commands}
\noindent \sloppy A typed scene, source, filter, transition, and output model serialized to the native OBS scene collection, allowing the agent to compose and control live production setups.

\bigskip
\begin{toollist}
\item[\famlabel{project}] \tool{project\_info} \tool{project\_json} \tool{project\_save}
\item[\famlabel{scene}] \tool{scene\_add} \tool{scene\_remove} \tool{scene\_duplicate} \tool{scene\_set\_active} \tool{scene\_list}
\item[\famlabel{source}] \tool{source\_add} \tool{source\_remove} \tool{source\_duplicate} \tool{source\_set} \tool{source\_transform} \tool{source\_list}
\item[\famlabel{filter}] \tool{filter\_add} \tool{filter\_remove} \tool{filter\_set} \tool{filter\_list} \tool{filter\_available}
\item[\famlabel{audio}] \tool{audio\_add} \tool{audio\_remove} \tool{audio\_volume} \tool{audio\_mute} \tool{audio\_unmute} \tool{audio\_monitor} \tool{audio\_list}
\item[\famlabel{transition}] \tool{transition\_add} \tool{transition\_remove} \tool{transition\_set\_active} \tool{transition\_duration} \tool{transition\_list}
\item[\famlabel{output}] \tool{output\_streaming} \tool{output\_recording} \tool{output\_settings} \tool{output\_info} \tool{output\_presets}
\end{toollist}

\subsection{LibreOffice Writer\,---\,30 commands}
\noindent \sloppy ODF-backed Writer operations over the live UNO bridge, implementing the Writer slice of the shared \texttt{cli-anything-libreoffice} harness for creating and editing rich text documents.

\bigskip
\begin{toollist}
\item[\famlabel{document}] \tool{document\_new} \tool{document\_open} \tool{document\_save} \tool{document\_info} \tool{document\_profiles} \tool{document\_json}
\item[\famlabel{writer}] \tool{writer\_add\_paragraph} \tool{writer\_add\_heading} \tool{writer\_add\_list} \tool{writer\_add\_table} \tool{writer\_table\_list} \tool{writer\_table\_insert\_row} \tool{writer\_table\_set\_cell} \tool{writer\_table\_set\_row\_background} \tool{writer\_add\_page\_break} \tool{writer\_remove} \tool{writer\_list} \tool{writer\_set\_text}
\item[\famlabel{style}] \tool{style\_create} \tool{style\_modify} \tool{style\_list} \tool{style\_apply} \tool{style\_remove}
\item[\famlabel{export}] \tool{export\_presets} \tool{export\_preset\_info} \tool{export\_render}
\item[\famlabel{session}] \tool{session\_status} \tool{session\_undo} \tool{session\_redo} \tool{session\_history}
\end{toollist}

\subsection{Kdenlive\,---\,29 commands}
\noindent \sloppy A project-file editor for Kdenlive's MLT XML format, using \texttt{melt} for rendering and writing a native \texttt{.kdenlive} project with timeline clips, filters, transitions, and guides.

\bigskip
\begin{toollist}
\item[\famlabel{project}] \tool{project\_info} \tool{project\_json} \tool{project\_save} \tool{project\_profiles}
\item[\famlabel{bin}] \tool{bin\_import} \tool{bin\_list} \tool{bin\_get}
\item[\famlabel{timeline}] \tool{timeline\_add\_track} \tool{timeline\_add\_clip} \tool{timeline\_remove\_clip} \tool{timeline\_trim} \tool{timeline\_split} \tool{timeline\_move} \tool{timeline\_list}
\item[\famlabel{filter}] \tool{filter\_add} \tool{filter\_set} \tool{filter\_list} \tool{filter\_available}
\item[\famlabel{transition}] \tool{transition\_add} \tool{transition\_set} \tool{transition\_list}
\item[\famlabel{guide}] \tool{guide\_add} \tool{guide\_list}
\item[\famlabel{export}] \tool{export\_xml} \tool{export\_presets}
\item[\famlabel{session}] \tool{session\_status} \tool{session\_undo} \tool{session\_redo} \tool{session\_history}
\end{toollist}

\subsection{Godot\,---\,26 commands}
\noindent \sloppy Godot project assets, scenes, scripts, and export configuration, accessed via the engine's conventions and CLI runtime for project creation, scene editing, and build export.

\bigskip
\begin{toollist}
\item[\famlabel{engine}] \tool{engine\_version} \tool{engine\_status}
\item[\famlabel{editor}] \tool{editor\_open}
\item[\famlabel{project}] \tool{project\_create} \tool{project\_info} \tool{project\_scenes} \tool{project\_scripts} \tool{project\_resources} \tool{project\_reimport} \tool{project\_set\_setting} \tool{project\_add\_input\_action} \tool{project\_add\_input\_key}
\item[\famlabel{scene}] \tool{scene\_create} \tool{scene\_read} \tool{scene\_add\_node} \tool{scene\_set\_property} \tool{scene\_set\_control\_layout}
\item[\famlabel{script}] \tool{script\_run} \tool{script\_inline} \tool{script\_validate} \tool{script\_read} \tool{script\_write} \tool{script\_append}
\item[\famlabel{export}] \tool{export\_presets} \tool{export\_build}
\item[\famlabel{session}] \tool{session}
\end{toollist}

\subsection{draw.io\,---\,25 commands}
\noindent \sloppy The native \texttt{.drawio} XML/mxGraph model, manipulated through a constrained diagram API that exposes pages, shapes, connectors, and export operations.

\bigskip
\begin{toollist}
\item[\famlabel{project}] \tool{project\_new} \tool{project\_open} \tool{project\_save} \tool{project\_info} \tool{project\_xml}
\item[\famlabel{page}] \tool{page\_add} \tool{page\_remove} \tool{page\_rename} \tool{page\_list}
\item[\famlabel{shape}] \tool{shape\_add} \tool{shape\_remove} \tool{shape\_list} \tool{shape\_label} \tool{shape\_move} \tool{shape\_resize} \tool{shape\_style}
\item[\famlabel{connect}] \tool{connect\_add} \tool{connect\_remove} \tool{connect\_label} \tool{connect\_style} \tool{connect\_list}
\item[\famlabel{export}] \tool{export\_render}
\item[\famlabel{session}] \tool{session\_undo} \tool{session\_redo} \tool{session\_status}
\end{toollist}

\subsection{Audacity\,---\,16 commands}
\noindent \sloppy A live client for Audacity's Mod-Script-Pipe, treating the running GUI project as the source of truth for importing audio, making selections, and applying effects.

\bigskip
\begin{toollist}
\item[\famlabel{import / track}] \tool{import} \tool{track\_new} \tool{get\_info}
\item[\famlabel{select}] \tool{select\_all} \tool{select\_time} \tool{select\_tracks}
\item[\famlabel{effects}] \tool{amplify} \tool{normalize} \tool{fade\_in} \tool{fade\_out} \tool{reverse} \tool{change\_pitch} \tool{change\_speed}
\item[\famlabel{label / export / raw}] \tool{label\_add} \tool{export} \tool{run}
\end{toollist}

\subsection{Chrome\,---\,15 commands}
\noindent \sloppy A restricted Chrome DOM surface (DOMShell), exposed through page, accessibility-tree, and action commands for navigation, file-system access, and element interaction.

\bigskip
\begin{toollist}
\item[\famlabel{page}] \tool{page\_open} \tool{page\_reload} \tool{page\_back} \tool{page\_forward} \tool{page\_info}
\item[\famlabel{fs}] \tool{fs\_ls} \tool{fs\_cd} \tool{fs\_cat} \tool{fs\_grep} \tool{fs\_pwd}
\item[\famlabel{act}] \tool{act\_click} \tool{act\_type}
\item[\famlabel{session}] \tool{session\_status} \tool{session\_daemon\_start} \tool{session\_daemon\_stop}
\end{toollist}

\subsection{LibreOffice Impress\,---\,14 commands}
\noindent \sloppy Native \texttt{.odp} slide, content, element, and export commands over ODF/UNO, implementing the Impress slice of the shared \texttt{cli-anything-libreoffice} harness.

\bigskip
\begin{toollist}
\item[\famlabel{document}] \tool{document\_save} \tool{document\_info}
\item[\famlabel{impress}] \tool{impress\_add\_slide} \tool{impress\_remove\_slide} \tool{impress\_set\_content} \tool{impress\_list\_slides} \tool{impress\_add\_element} \tool{impress\_remove\_element} \tool{impress\_move\_slide} \tool{impress\_duplicate\_slide} \tool{impress\_get\_slide}
\item[\famlabel{export}] \tool{export\_presets} \tool{export\_preset\_info} \tool{export\_render}
\end{toollist}

\subsection{VS Code\,---\,13 commands}
\noindent \sloppy The installed \texttt{code} binary, invoked to open paths, diff and merge files, manage extensions, and access settings and keybindings.

\bigskip
\begin{toollist}
\item[\famlabel{open}] \tool{open} \tool{open\_new\_window} \tool{open\_user\_settings\_json} \tool{open\_keybindings\_json}
\item[\famlabel{navigate}] \tool{goto} \tool{diff} \tool{merge}
\item[\famlabel{workspace}] \tool{add\_folder} \tool{wait\_file\_closed}
\item[\famlabel{extensions}] \tool{install\_extension} \tool{uninstall\_extension} \tool{list\_extensions}
\item[\famlabel{status}] \tool{status}
\end{toollist}

\subsection{VLC\,---\,13 commands}
\noindent \sloppy Direct \texttt{vlc} and \texttt{cvlc} invocations to launch media, open files at specific times or segments, capture snapshots, transcode audio and video, and manage configuration.

\bigskip
\begin{toollist}
\item[\famlabel{app}] \tool{app\_open} \tool{app\_help}
\item[\famlabel{open}] \tool{open\_media\_file} \tool{open\_media\_at\_time} \tool{open\_network\_stream} \tool{open\_preferences\_file}
\item[\famlabel{play}] \tool{play\_fullscreen} \tool{play\_paused\_segment}
\item[\famlabel{capture}] \tool{snapshot\_frame}
\item[\famlabel{convert}] \tool{convert\_audio\_mp3} \tool{convert\_audio\_wav} \tool{convert\_video\_mp4}
\item[\famlabel{config}] \tool{reset\_user\_config}
\end{toollist}

\subsection{LibreOffice Calc\,---\,10 commands}
\noindent \sloppy Bounded sheet and cell operations over ODF/UNO, implementing the Calc slice of the shared \texttt{cli-anything-libreoffice} harness for spreadsheet creation and manipulation.

\bigskip
\begin{toollist}
\item[\famlabel{document}] \tool{document\_save} \tool{document\_info}
\item[\famlabel{sheet}] \tool{calc\_list\_sheets} \tool{calc\_add\_sheet} \tool{calc\_rename\_sheet}
\item[\famlabel{cell}] \tool{calc\_get\_cell} \tool{calc\_set\_cell}
\item[\famlabel{export}] \tool{export\_presets} \tool{export\_preset\_info} \tool{export\_render}
\end{toollist}

\subsection{Thunderbird\,---\,10 commands}
\noindent \sloppy The installed executable's profile-aware launch and compose capabilities, with every command naming the benchmark profile to open mail, address book, calendar, and compose emails.

\bigskip
\begin{toollist}
\item[\famlabel{app}] \tool{app\_open} \tool{app\_help}
\item[\famlabel{open}] \tool{open\_mail} \tool{open\_addressbook} \tool{open\_calendar}
\item[\famlabel{compose}] \tool{compose\_email} \tool{compose\_cc\_bcc} \tool{compose\_attachment} \tool{mailto\_email}
\item[\famlabel{desktop handler}] \tool{xdg\_email}
\end{toollist}

\section{Task-Weave}

\subsection{Seed Examples}
\label{app:seeds}
\begin{figure*}[t]
\centering
\begin{minipage}{0.32\textwidth}
  \centering
  \includegraphics[width=\linewidth]{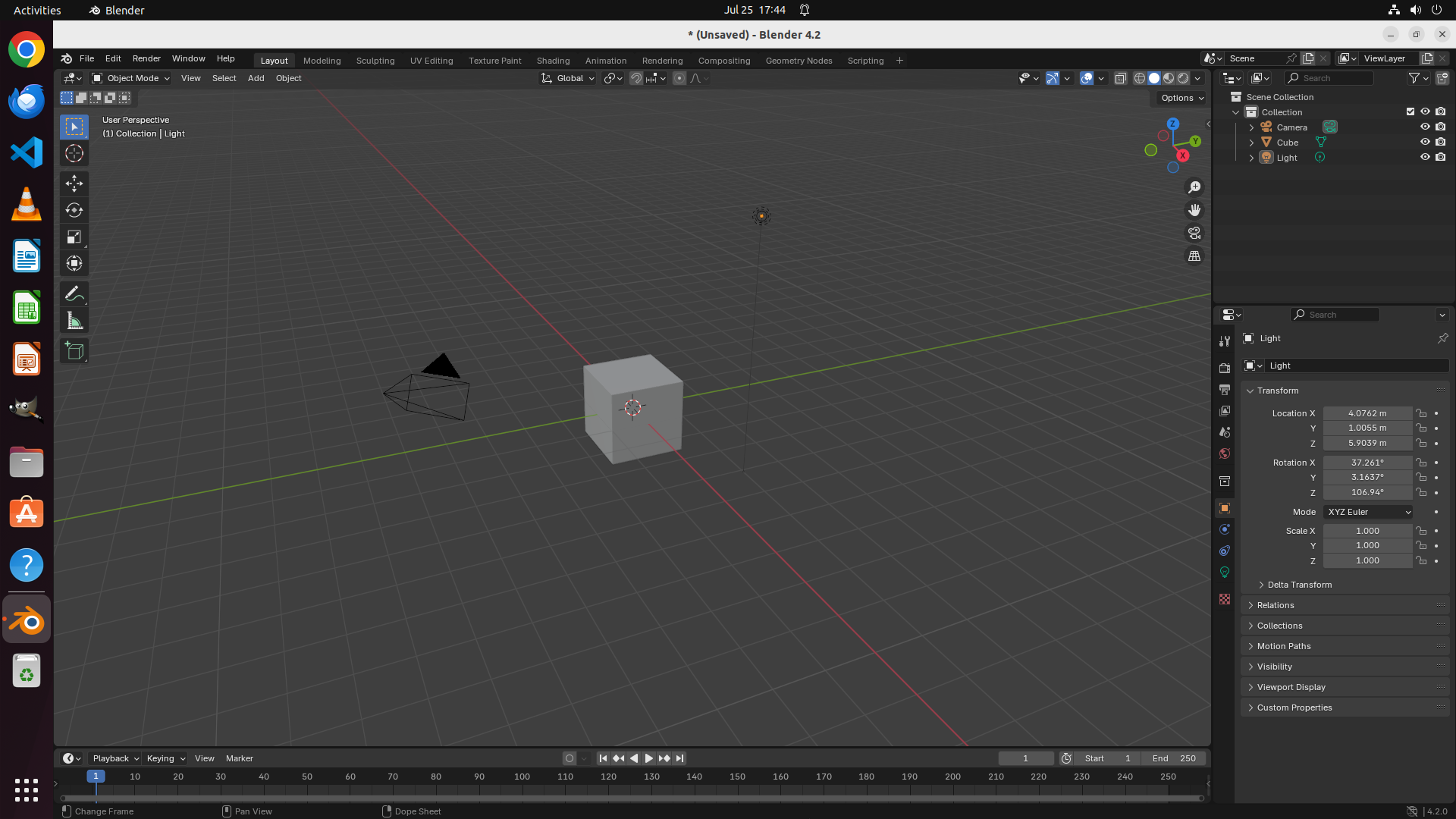}
  \caption*{\textbf{Blender.} This Objaverse product seed opens a concrete webcam-cover 3D asset, grounding modeling tasks in real mesh and material structure.}
\end{minipage}\hfill
\begin{minipage}{0.32\textwidth}
  \centering
  \includegraphics[width=\linewidth]{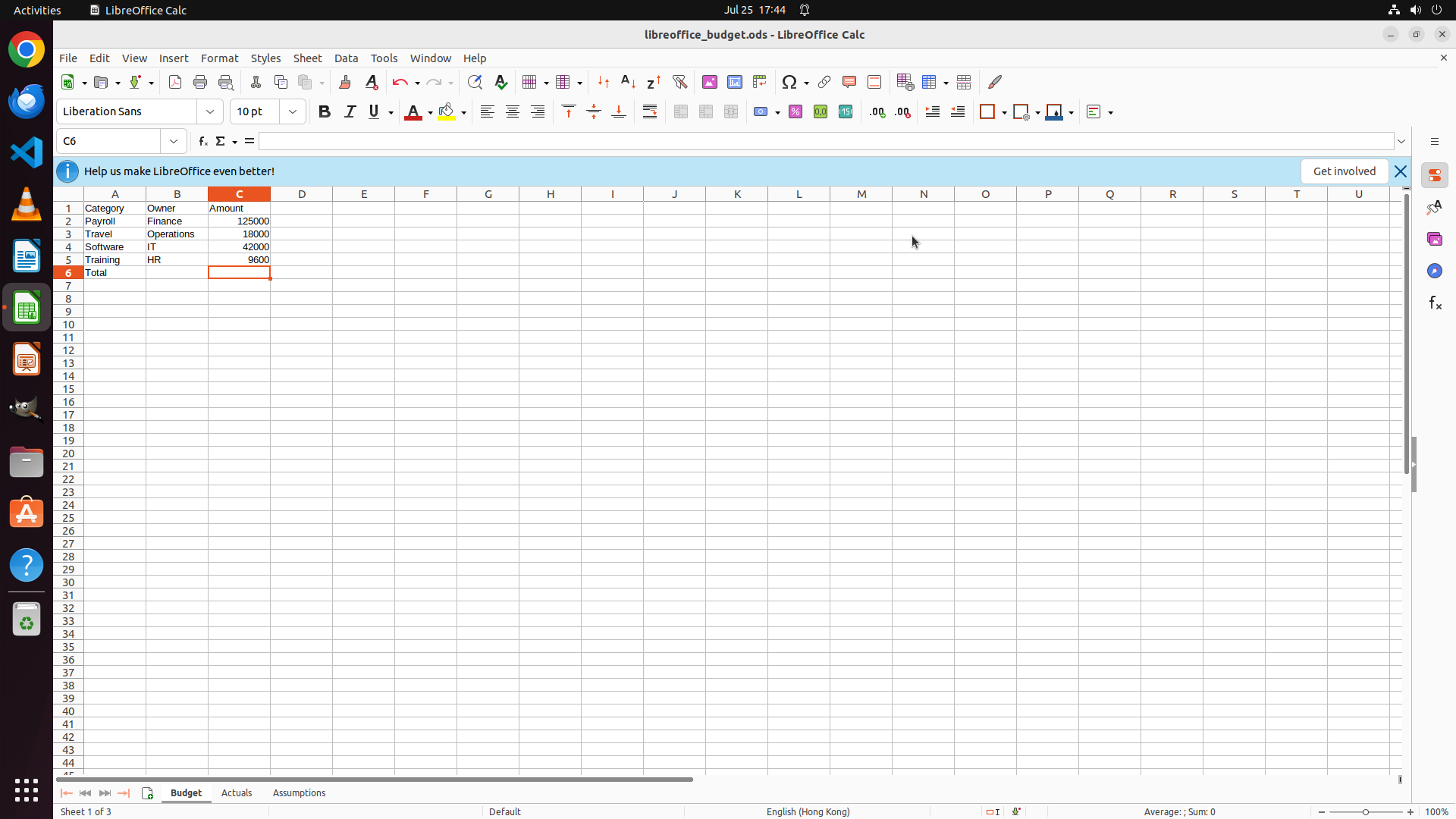}
  \caption*{\textbf{LibreOffice Calc.} This budget workbook seed provides real sheets, headers, rows, and editable cells for grounded spreadsheet tasks.}
\end{minipage}\hfill
\begin{minipage}{0.32\textwidth}
  \centering
  \includegraphics[width=\linewidth]{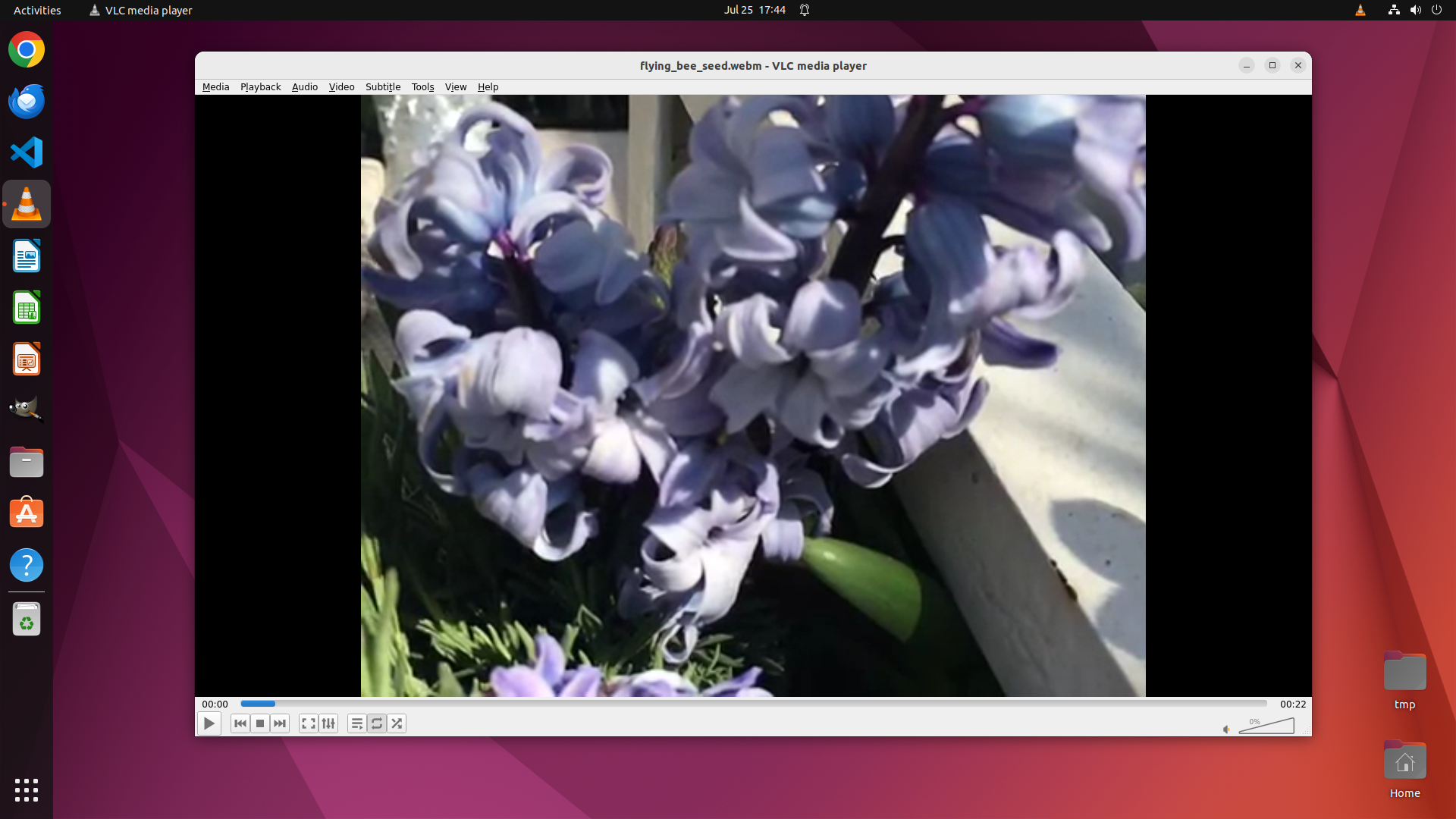}
  \caption*{\textbf{VLC.} This media seed opens a concrete video file, grounding playback and media-control tasks in a real duration and visual stream.}
\end{minipage}

\vspace{0.8em}

\begin{minipage}{0.32\textwidth}
  \centering
  \includegraphics[width=\linewidth]{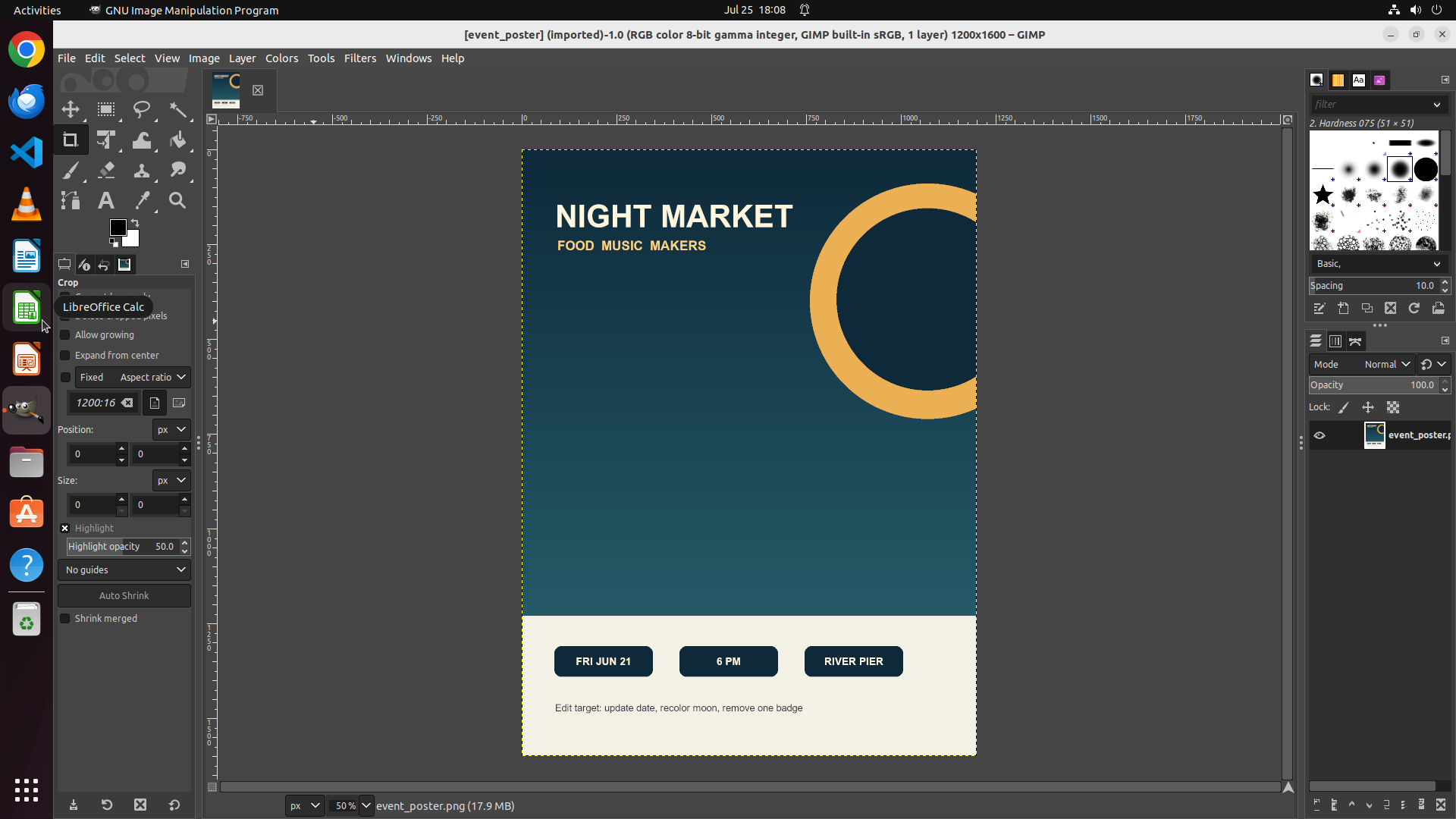}
  \caption*{\textbf{GIMP.} This poster seed exposes a real raster design with stable regions for annotation, cropping, banner, and export tasks.}
\end{minipage}\hfill
\begin{minipage}{0.32\textwidth}
  \centering
  \includegraphics[width=\linewidth]{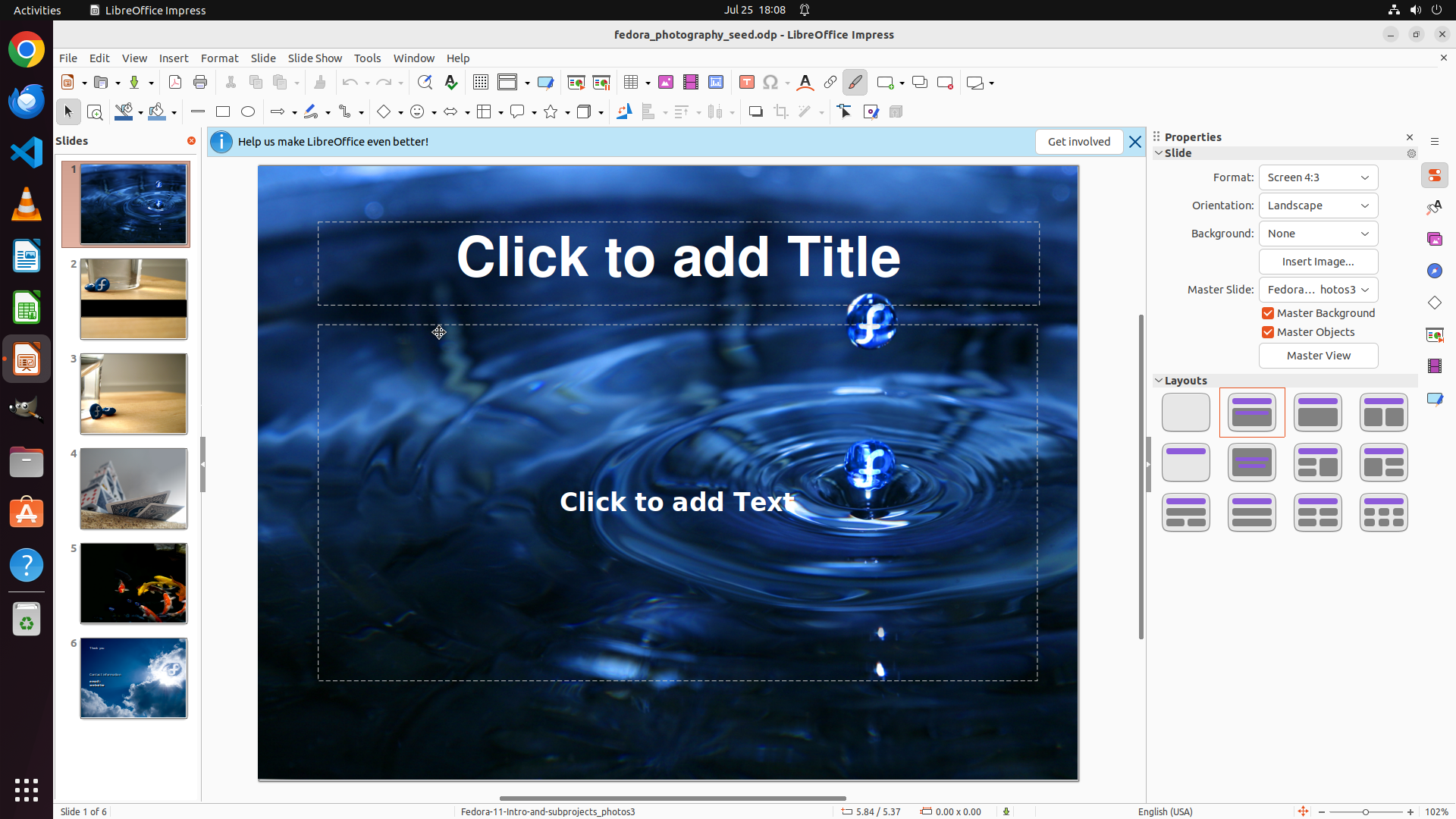}
  \caption*{\textbf{LibreOffice Impress.} This photo-heavy deck seed grounds presentation tasks in existing slides, layouts, images, and editable text frames.}
\end{minipage}\hfill
\begin{minipage}{0.32\textwidth}
  \centering
  \includegraphics[width=\linewidth]{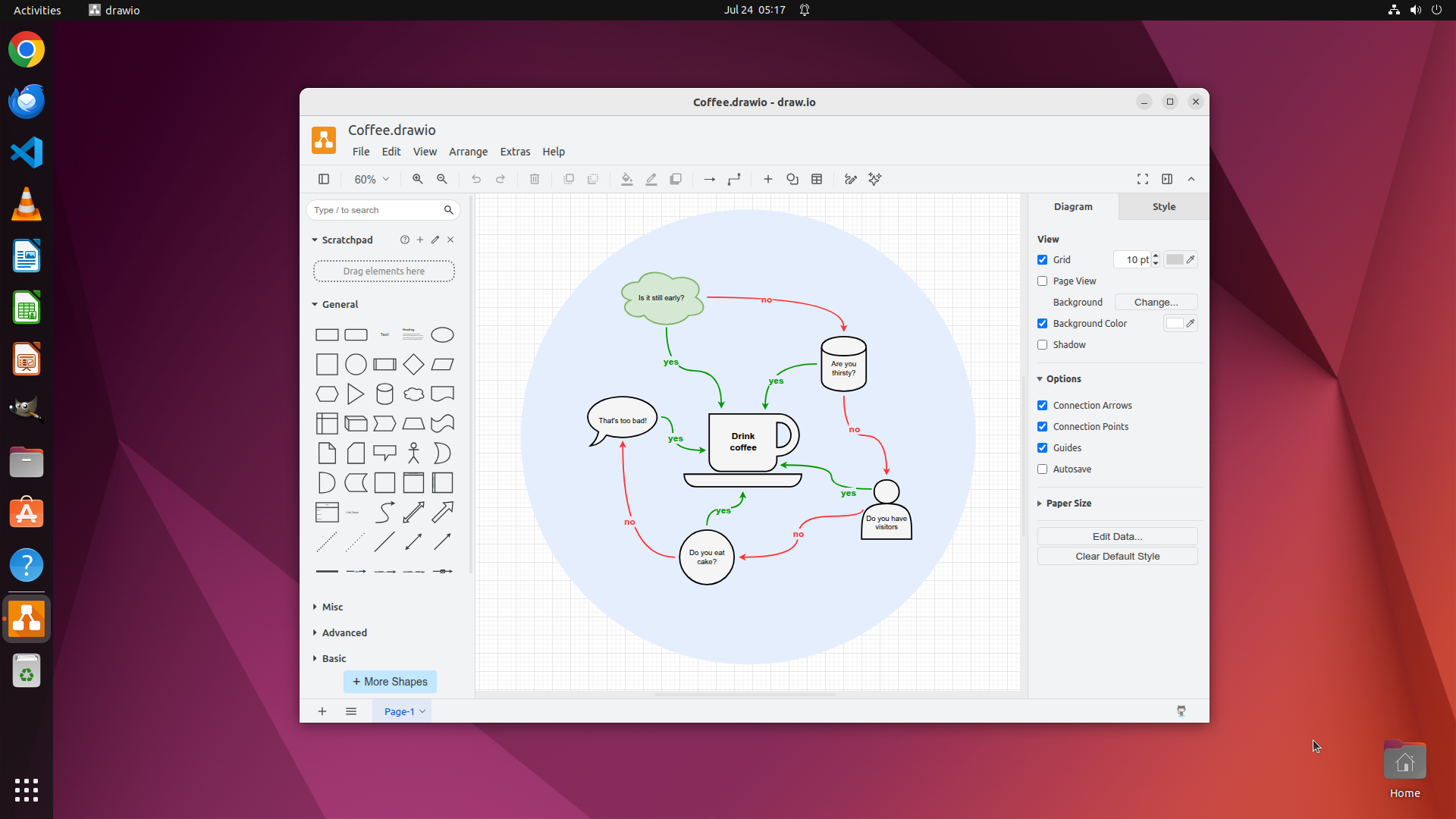}
  \caption*{\textbf{Draw.io.} This flowchart seed grounds diagram-editing tasks in existing nodes, connectors, labels, and saved \texttt{.drawio} XML.}
\end{minipage}

\caption{Six opened seed examples used to ground task synthesis across diverse GUI applications. Each seed is a concrete project or file with metadata and verifiable app state, rather than a generic natural-language prompt.}
\label{fig:appendix-seed-final-six}
\end{figure*}

A \emph{seed} is a concrete, pre-loaded application state---an actual project, document, or media file opened in its native GUI app---that anchors task synthesis in real, verifiable context rather than a generic natural-language prompt. Each seed carries genuine content and metadata (e.g., a populated spreadsheet, a raster design with stable regions, or a saved \texttt{.drawio} graph), so synthesized tasks reference elements that truly exist and produce app state that can be checked deterministically. Grounding generation in seeds reduces hallucinated targets, yields more executable and diverse instructions, and makes success criteria objective. Figure~\ref{fig:appendix-seed-final-six} shows six representative seeds spanning 3D modeling, spreadsheets, media playback, image editing, presentations, and diagramming; together they illustrate the breadth of GUI applications and file types our pipeline builds on.

\noindent Seeds are not hand-authored. A coding agent (Codex) searches for and downloads real, diverse source files from public repositories and asset libraries, then uses format-conversion tools to normalize each one into an agent-readable, synthesis-friendly project state---so the seed pool scales automatically while staying grounded in authentic, heterogeneous content rather than templated fixtures.

\section{Steer-Path Rollouts}
\subsection{Rollout Agent Prompt}
\label{app:rollout-message}
Figure~\ref{fig:rollout-message} shows the message format of a data-generation rollout (the Kimi~K2.5 backbone on a draw.io task). Two properties are worth noting. First, the rollout interface exposes GUI and CLI \emph{jointly}: the system prompt lists the application's CLI tool registry alongside \texttt{pyautogui}, \texttt{wait}, and \texttt{terminate}, and the baseline modality guidance is the generic instruction to ``use CLI for precise operations and GUI for visual tasks.'' Second, this is where Path-Steer enters: the task guidance field carries the efficiency-aware prior for this attempt (an ordered plan such as ``First \dots''), steering the backbone toward a shorter hybrid path during data generation. The agent then interleaves CLI edits with GUI waits, and each CLI call returns a structured JSON result (\texttt{rc}, \texttt{output}) that becomes the next observation. Path-Steer is used \emph{only} during these rollouts; at evaluation the guidance field is empty (Appendix~\ref{app:actor-prompt}). The task text is elided below; the point is the interface and the guided GUI+CLI interleaving, not the specific diagram.

\subsection{Steer-Path Rollout Examples}
\label{app:steer-path}

In this VLC desktop task, the \emph{w/ Path-Steer} trajectory overall outperformed the \emph{w/o Path-Steer} trajectory. Although the \emph{w/ Path-Steer} trajectory made several ineffective attempts with VLC CLI parameters in the early stage, it was able to switch to a feasible alternative solution in time, namely using ffmpeg to complete the video rotation, and eventually succeeded in both exporting the corrected video and opening the Audio Effects panel. In contrast, while the \emph{w/o Path-Steer} trajectory followed a more intuitive GUI-based workflow, it became stuck in repeated searching and ineffective interactions during the filter configuration stage, and ultimately failed to complete the task. This case suggests that Path-Steer may not necessarily reduce local trial-and-error, but it can significantly improve task completion, error recovery, and goal convergence in complex desktop environments.

\begin{figure*}[t]
\centering
\includegraphics[width=1\linewidth]{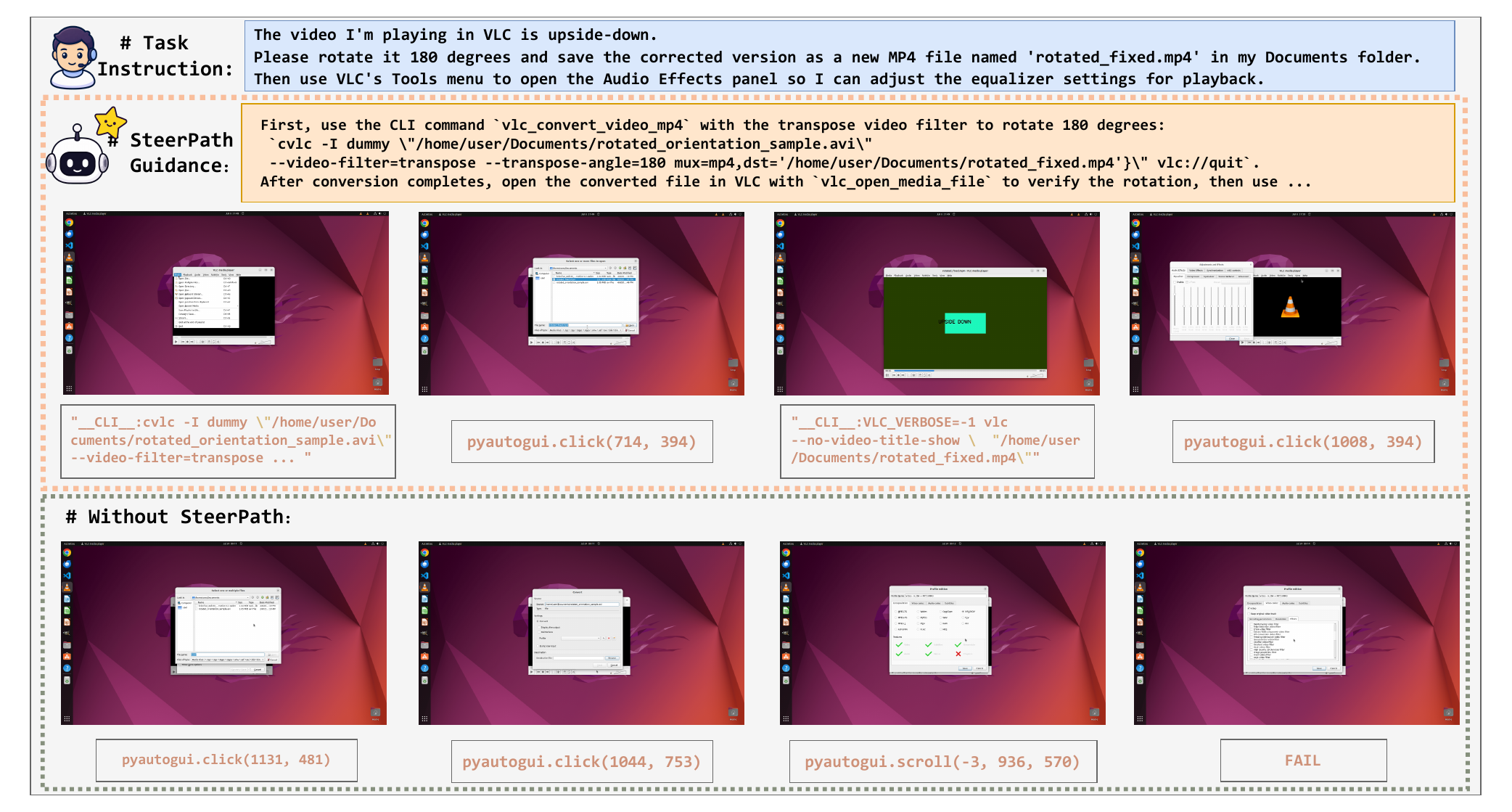}

\caption{VLC task trajectories \emph{w/ Path-Steer} and \emph{w/o Path-Steer}. The Path-Steer trajectory shows optimal solution using CLI tools and GUI actions, whereas the trajectory without sterr-path guidance becomes stuck during filter configuration and fails the task.}
\label{fig:vlc-steer-path}
\end{figure*}

\section{VLM Judge}
\label{app:vlm-judge}

We score task success with a VLM-as-judge, following the now-standard practice of using strong (vision-)language models as automatic evaluators of agent trajectories; in the GUI and computer-use setting in particular, VLM judges are widely used both to score task completion from screenshots and to curate training trajectories for post-training. Independent of the task synthesizer and of any rollout reward, the judge takes the task instruction, a compact rendering of the trajectory, and a chronological sample of screenshots, and returns a score in $[0,1]$ with a short justification. Its rubric treats CLI output and exported-artifact evidence as authoritative for file-producing tasks while using the final screenshot as primary evidence otherwise. We run the judge (GPT-5.4~\cite{openai2026gpt54}) at temperature $0.1$ with up to $15$ screenshots; the full prompt is shown in Figure~\ref{fig:judge-prompt}.

\subsection{Agreement with Human Labels}
Because Score is our headline CUA-Verse metric and the same judge filters training data, we validate it against human judgment. Three annotators independently re-scored a stratified sample of trajectories for all 16 applications, \emph{blind} to the judge's score. For each application we drew 30 trajectories the judge \emph{accepted} ($\text{score}\ge0.75$) and 30 it \emph{rejected} ($\text{score}<0.75$), and labeled each as fully success, partial success, or failure. We take the binary decision positive $=$ human ``fully success'' and negative $=$ otherwise, matching the semantics of the $0.75$ acceptance threshold used to filter training data. Because the sample is stratified rather than proportional, we reweight each application's confusion cells by its true acceptance rate $p$ (Table~\ref{tab:judge-human}, second column) before computing the aggregate agreement and $\kappa$.

\noindent Two findings anchor the judge's reliability. On the \textbf{acceptance set} (480 trajectories) the judge attains \textbf{99.0\% precision} (475/480), and crucially \emph{not one} accepted trajectory was a human ``failure''---the five imperfections are all partial successes---so the label-noise upper bound on training data is $\le\!1.0\%$. On the \textbf{rejection set} (480 trajectories) the false-negative rate is \textbf{5.2\%} (25 fully-successful trajectories scored low); these cost data yield but do not inflate reported Score. Reweighted to the true distribution, judge--human agreement is \textbf{97.0\%} and \textbf{Cohen's $\kappa=0.94$} (Landis--Koch ``almost perfect'').

\noindent Under a lenient criterion that counts partial successes as positive, the picture is unchanged where it matters: the judge accepts \emph{zero} true failures (precision $100\%$; all 480 accepted trajectories are at least partial successes). Its rejection set is dominated by partial successes ($406/480$) with only $49$ true failures, confirming that the $0.75$ threshold deliberately screens out partially-completed work---the intended behavior for a training-data filter---rather than confusing success with failure.

\begin{table}[t]
\centering
\small
\setlength{\tabcolsep}{4.5pt}
\renewcommand{\arraystretch}{1.1}
\begin{tabular}{lccccc}
\toprule
App & $p$ & FP & FN & Prec.\,(\%) & FNR\,(\%) \\
\midrule
\rowcolor{OSWorldBlue}Chrome            & 0.53 & 0 & 3 & 100.0 & 10.0 \\
\rowcolor{OSWorldBlue}GIMP              & 0.52 & 1 & 3 & 96.7  & 10.0 \\
\rowcolor{OSWorldBlue}LibreOffice Calc  & 0.56 & 0 & 1 & 100.0 & 3.3  \\
\rowcolor{OSWorldBlue}LibreOffice Impress&0.51 & 0 & 2 & 100.0 & 6.7  \\
\rowcolor{OSWorldBlue}LibreOffice Writer& 0.53 & 0 & 0 & 100.0 & 0.0  \\
\rowcolor{OSWorldBlue}Thunderbird       & 0.59 & 0 & 0 & 100.0 & 0.0  \\
\rowcolor{OSWorldBlue}VLC               & 0.57 & 0 & 0 & 100.0 & 0.0  \\
\rowcolor{OSWorldBlue}VS Code           & 0.52 & 0 & 2 & 100.0 & 6.7  \\
\rowcolor{CUAOrange}Audacity            & 0.52 & 0 & 0 & 100.0 & 0.0  \\
\rowcolor{CUAOrange}Blender             & 0.49 & 0 & 2 & 100.0 & 6.7  \\
\rowcolor{CUAOrange}Draw.io             & 0.48 & 0 & 3 & 100.0 & 10.0 \\
\rowcolor{CUAOrange}Godot               & 0.46 & 2 & 2 & 93.3  & 6.7  \\
\rowcolor{CUAOrange}Kdenlive            & 0.59 & 1 & 1 & 96.7  & 3.3  \\
\rowcolor{CUAOrange}OBS                 & 0.54 & 1 & 3 & 96.7  & 10.0 \\
\rowcolor{CUAOrange}QGIS                & 0.57 & 0 & 2 & 100.0 & 6.7  \\
\rowcolor{CUAOrange}Zotero              & 0.48 & 0 & 1 & 100.0 & 3.3  \\
\midrule
\textbf{Overall} & — & \textbf{5} & \textbf{25} & \textbf{99.0} & \textbf{5.2} \\
\bottomrule
\end{tabular}
\caption{\textbf{VLM judge vs.\ human labels, all 16 applications.} Three annotators, blind to the judge score, labeled 30 judge-accepted and 30 judge-rejected trajectories per application. $p$ is the true acceptance rate; FP counts accepted trajectories that were not a human ``fully success'' (out of 30, all of them partial successes---zero hard failures); FN counts rejected trajectories that were in fact fully successful (out of 30). Precision is on the acceptance set and FNR on the rejection set. Reweighting each application by $p$ gives an overall agreement of 97.0\% and Cohen's $\kappa=0.94$. Blue rows are OSWorld applications; orange rows are CUA-Universe extensions.}
\label{tab:judge-human}
\end{table}

\section{Rollout Harness and Budget}
\label{app:compute}

\paragraph{Rollout Harness and hardware.} Each task runs in an isolated OSWorld virtual machine provisioned with 4~vCPU and 4\,GB of guest RAM on a copy-on-write overlay disk, with a 3\,s post-action settle before each observation. All agents share a 60-step budget; observation history follows each baseline's default, while our Qwen-family models use a 3-frame history with at most 4 images per step.

\paragraph{Rollout budget.} Trajectory collection dominates the wall-clock cost of building the fine-tuning dateset, because each rollout drives a full GUI environment rather than a single forward pass. A single rollout takes on average $\bar{t}\approx5$ minutes end-to-end (environment reset, per-step model calls, action execution, and the post-action settle), and a training run consumes on the order of $N\approx10^4$ rollouts. Because each environment needs only $\sim$4 vCPU and no GPU, this collection is CPU-bound and embarrassingly parallel. Concretely, a single commodity \textbf{128-core CPU server} hosts $P=\lfloor 128/4\rfloor=32$ environments concurrently, so the full $N\bar{t}\approx50{,}000$ VM-minutes reduce to $N\bar{t}/P\approx1{,}560$ minutes---just over \textbf{one day} of wall-clock time on that one machine. This is the key practicality of our recipe: the entire data-generation pipeline that produces our 9B model fits on a single ordinary CPU box, with no GPU cluster and no multi-machine orchestration. Throughput in practice is slightly below this ceiling because of environment-reset overhead and occasional VM stalls (we observe $\sim$1.3--1.5 calendar days on one 128-core host), and the job is trivially shardable across additional machines when faster turnaround is needed.

\section{Training Details}
\label{app:training}

\subsection{Data}
All records satisfy VLM score $\geq 0.75$ and are step-level examples in Qwen XML format with a three-step history context. The CUA-Verse pool is balanced at 17{,}511 records per app across Audacity, Blender, draw.io, Godot, Kdenlive, OBS, QGIS, and Zotero. The OSWorld pool is balanced at 11{,}915 records per app across LibreOffice Writer, LibreOffice Calc, LibreOffice Impress, VS Code, VLC, Thunderbird, GIMP, and Chrome. The OSWorld build enforces exactly four images per record; the CUA-Verse build is VLM-filtered but not strictly four-image (22{,}590 records have an image count other than four). Figure~\ref{fig:data-composition} summarizes the overall split and the per-application episode counts for both pools, and Table~\ref{tab:data} reports the final split sizes.

\begin{table}[t]
\centering
\begin{tabular}{lrrr}
\toprule
Split & Episodes & Step records & Total (GB) \\
\midrule
CUA-Verse & 2{,}526 & 140{,}088 & 172.97 \\
OSWorld   & 2{,}397 & 95{,}320 & 113.67 \\
\midrule
Total     & 4{,}923 & 235{,}408 & 286.64 \\
\bottomrule
\end{tabular}
\caption{Final training splits. ``Total (GB)'' is the on-disk size including all screenshots; step records reference images by path.}
\label{tab:data}
\end{table}

\subsection{Base Model and Fine-tuning}
We use \texttt{Qwen3.5-9B} as the base model in bfloat16. Fine-tuning is performed with LoRA while the base weights, vision tower, and multimodal aligner remain frozen; only the language-model linear modules  are trainable. Table~\ref{tab:optim} lists the optimization and sequence settings used for fine-tuning.

\begin{table}[h]
\centering
\begin{tabular}{lr}
\toprule
Parameter & Value \\
\midrule
Tuner type & LoRA \\
LoRA rank & 8 \\
LoRA $\alpha$ & 32 \\
LoRA dropout & 0.05 \\
LoRA bias & none \\
Target modules & all-linear \\
Vision tower & frozen \\
Multimodal aligner & frozen \\
\bottomrule
\end{tabular}
\caption{LoRA configuration.}
\label{tab:lora}
\end{table}

\begin{table}
\centering
\begin{tabular}{lr}
\toprule
Parameter & Value \\
\midrule
Optimizer & ms-swift default \\
Learning rate & $1\times10^{-4}$ \\
Warmup ratio & 0.05 \\
Per-device train batch size & 2 \\
Gradient accumulation & 1 \\
Effective batch size (8 GPUs) & 16 \\
Epoch argument & 3 \\
Max length & 18{,}000 \\
Max pixels & 602{,}112 \\
Image max token budget & 1{,}024 \\
Attention & Flash Attention 2 \\
DeepSpeed & ZeRO-2 \\
\bottomrule
\end{tabular}
\caption{Optimization and sequence settings.}
\label{tab:optim}
\end{table}

\begin{figure*}[!t]
  \centering
  \includegraphics[width=\linewidth]{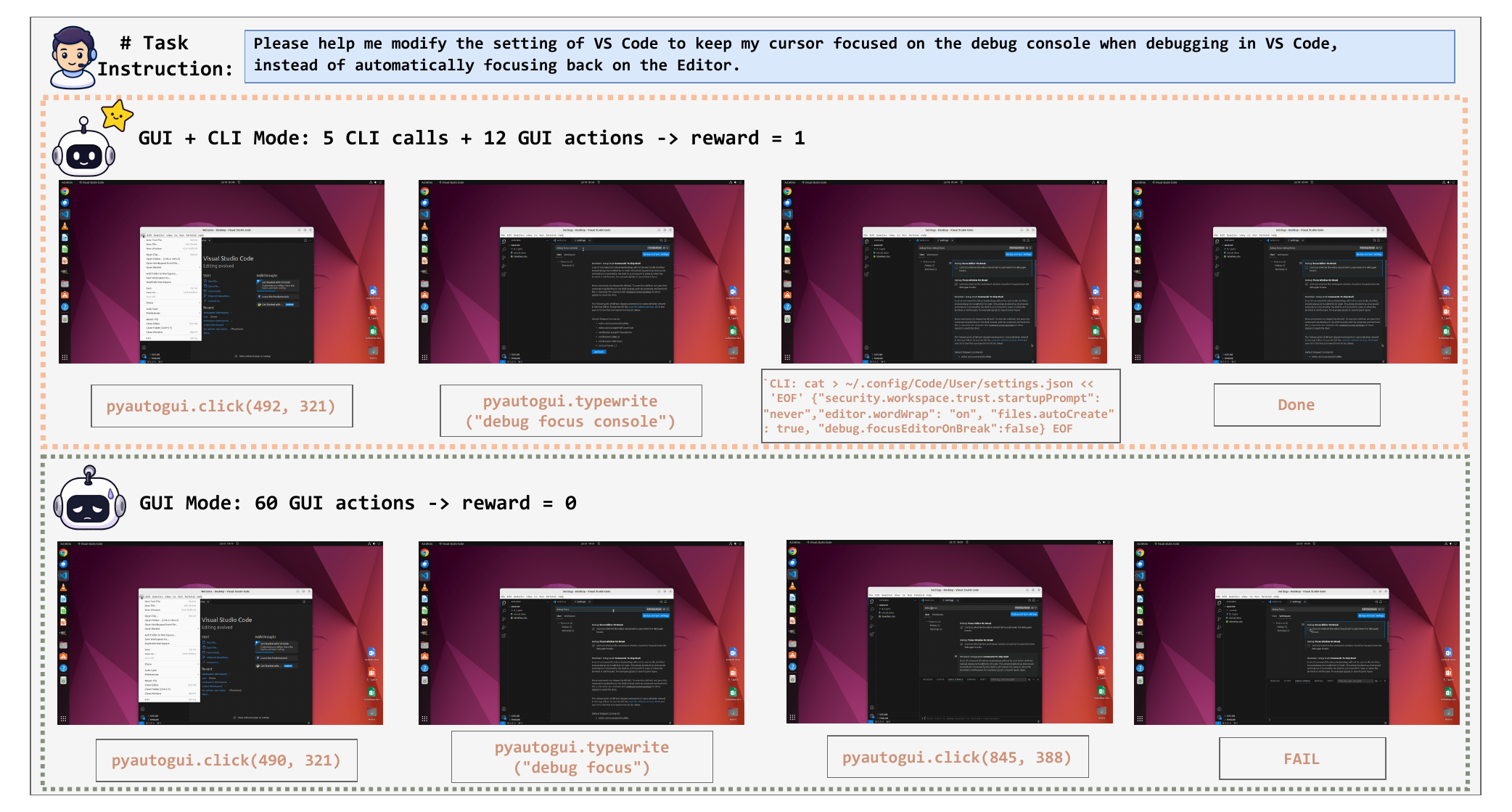}
  \caption{\textbf{An OSWorld VS~Code task under GUI+CLI vs.\ GUI-only interfaces.} \emph{GUI+CLI} mode combines GUI navigation with direct settings-file editing: it locates the target setting, writes \texttt{debug.focusEditorOnBreak: false} to \texttt{settings.json}, and verifies persistence, completing the task in 5 CLI calls and 12 GUI actions (reward~$1$). \emph{GUI-only} mode repeatedly searches and clicks through the settings UI but fails to commit the correct configuration, exhausting the 60-action budget (reward~$0$).}
  \label{fig:osworld-vscode}
\end{figure*}

\section{Experiments}
\subsection{Evaluation Agent Prompt}
\label{app:actor-prompt}
Each evaluated backbone uses its own actor prompt, and each is run in two modes across OSWorld and CUA-Verse: a CLI mode that additionally exposes the \texttt{execute\_cli} function, and a GUI mode that follows the standard OSWorld screenshot-and-\texttt{pyautogui} protocol. Figure~\ref{fig:actor-prompt} shows the Qwen CLI-mode prompt as a representative example. Regardless of backbone or mode, the path-hint field defaults to \texttt{none} at evaluation, so no task hint or solution guidance is appended and the evaluated policy is prior-free---the Path-Steer priors of Appendix~\ref{app:rollout-message} act only during data-generation rollouts.

\subsection{CUA-Verse Benchmark}
\label{app:cuaverse}
\begin{figure}[h]
\centering
\includegraphics[width=\linewidth]{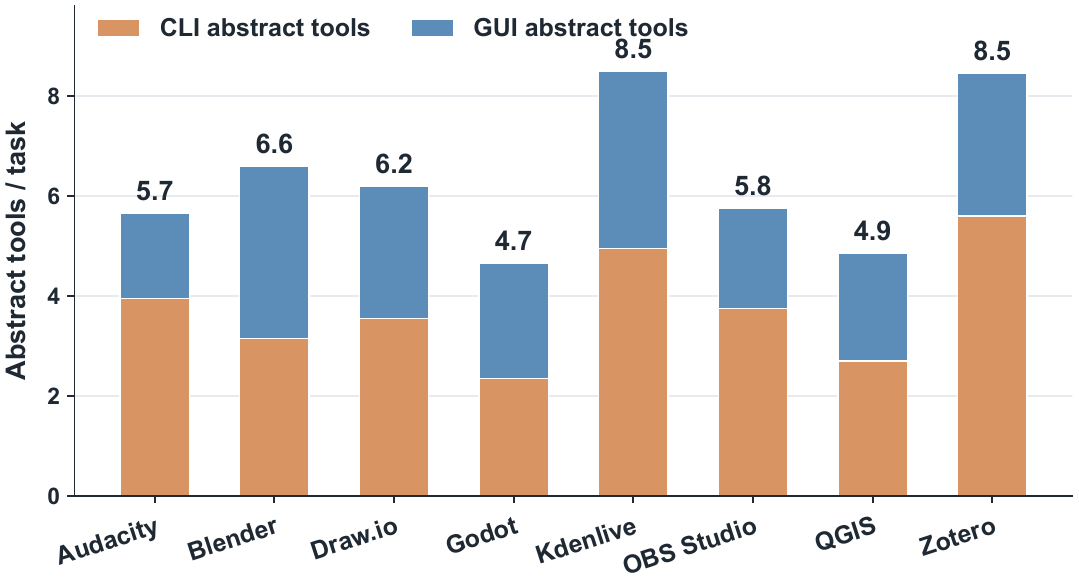}
\caption{\textbf{CUA-Verse task composition.} Mean number of abstract tools per task for each application, split into CLI (orange) and GUI (blue) operations. Every application requires both modalities.}
\label{fig:cuaverse-dist}
\end{figure}

CUA-Verse comprises 160 hybrid GUI+CLI tasks (eight applications $\times$ 20 tasks) synthesized by the same pipeline as the training data. To characterize what the tasks demand, we abstract each task's reference solution into a deduplicated set of high-level tools and classify each as GUI (normalized \texttt{keyboard}/\texttt{click}/\texttt{move}/\texttt{drag}/\texttt{scroll}/\texttt{vision}) or CLI (application command groups such as \texttt{godot\_scene\_*} or \texttt{qgis\_layer\_*}). Figure~\ref{fig:cuaverse-dist} reports the resulting per-application composition. Tasks require $5.7$ abstract tools on average, of which $59\%$ are CLI, and every application mixes both modalities---confirming that CUA-Verse is genuinely hybrid rather than solvable by either modality alone. The two extremes are illustrative: Blender is GUI-dominant (spatial 3D manipulation), whereas Zotero and Kdenlive are CLI-heavy (structured library and timeline edits). The eight applications are \emph{in-domain} by design, since we want to measure hybrid competence on the software the pipeline covers.

\section{OSWorld Example}
\label{app:osworld-example}
Figure~\ref{fig:osworld-vscode} contrasts the \emph{GUI+CLI} and \emph{GUI-only} interfaces on a single OSWorld task---modifying a VS~Code setting so the cursor stays focused on the debug console during debugging rather than snapping back to the editor. In \emph{GUI+CLI} mode the agent uses a few GUI actions to locate the relevant setting, then drops to the CLI to write \texttt{"debug.focusEditorOnBreak": false} directly into \texttt{settings.json} and verify that it persists, resolving the task in 5 CLI calls and 12 GUI actions (reward~$1$). In \emph{GUI-only} mode it repeatedly searches and clicks through the settings UI but never commits the correct configuration, exhausting the full 60-action budget (reward~$0$). The case makes the orchestration advantage concrete: the GUI \emph{locates} the setting, while the CLI performs the \emph{precise, verifiable write} that the GUI-only agent cannot reliably land.

\section{Limitations}
\label{app:limitations}
Our study leaves several directions open. First, CUA-Universe synthesizes single-application tasks: each task is grounded in one application's shared GUI+CLI state, which is what lets us build verifiable environments and steer efficient hybrid paths at scale. Cross-application workflows, where state is carried across several applications, are a natural extension of the same pipeline rather than a different design, and we leave them to future work. Second, we use the harvested trajectories for supervised fine-tuning only, so the learned policy is bounded by its data-generation backbone; because the pipeline already produces per-task verifiers, using them as rewards for reinforcement learning is a direct next step toward surpassing the teacher. Third, task success is scored by a VLM judge rather than per-task programmatic checks, which may introduce label noise; the judge--human validation in Appendix~\ref{app:vlm-judge} bounds this ($99.0\%$ acceptance precision, Cohen's $\kappa=0.94$), and we further mitigate it with a conservative acceptance threshold and by grounding the judge in CLI and exported-artifact evidence, but we do not eliminate it. Finally, application adaptation assumes software that is open-source or scriptable enough to expose a command-line surface, and our environments target desktop Linux; the coding-agent-driven construction is not tied to these choices in principle, but broadening to closed-source or non-desktop platforms remains open.

% Using the \centering command instead of \begin{center} ... \end{center} will save space
% Positioning your figure at the top of the page will save space and make the paper more readable
% Using 0.95\columnwidth in conjunction with the

\begin{figure*}[t]
\centering
\begin{lstlisting}[style=prompt]
[SYSTEM]
You are a computer use agent. You can operate the computer through both GUI
(mouse/keyboard) and CLI (command-line) tools. You are given a task instruction,
a screenshot of the current screen, and your previous interactions. Complete the
task by issuing one action per step. Choose the most efficient tool for each
step - use CLI for precise operations and GUI for visual tasks.
For each step, respond in EXACTLY this format:
{thought}
## Action:
{action}
## Code:
{code}
In the code section, use ONE of:
- A python block with pyautogui code (GUI action)
- A cli block with a shell command (CLI action, only when CLI tools are listed)
- A special function in a code block:
  - {"name": "computer.wait",      "parameters": {"time": 3}}
  - {"name": "computer.terminate", "parameters": {"status": "success"}}
  - {"name": "computer.terminate", "parameters": {"status": "failure"}}
CLI Tool: Execute a CLI tool using a cli block.
  - CLI modifies the file on disk. The GUI may not auto-update.
  - Always check CLI output (rc and stdout). If rc != 0, diagnose and retry.
  - You can chain multiple CLI steps before switching to GUI.
## Available CLI Tools  (draw.io registry, abridged)
drawio_project_new    | Create a new empty draw.io diagram file
drawio_project_save   | Save the current diagram to disk
drawio_shape_add      | Add a shape at a position (cylinder|rectangle|...)
drawio_shape_style    | Set a style property on a shape (fillColor, ...)
drawio_connect_add    | Add a labeled connector between two shapes
drawio_export_render  | Export the diagram to PNG/PDF/SVG
...   (full registry of project/shape/connect/page/export/session commands)
[Task guidance for this attempt: First ...]
--- Interaction history (elided) ---
[USER]  Instruction: {task_instruction}
        Previous actions: {action_history}
        Previous CLI results: {cli_result_history}
        {current_screenshot}
[ASSISTANT] {thought}
            ## Action: {next_action}
            ## Code: {python_or_cli_block}
... (this GUI/CLI interleaved turn repeats until terminate)
\end{lstlisting}
\caption{Message format of a data-generation rollout (Kimi~K2.5 on a draw.io task). The system prompt exposes GUI and CLI jointly with generic modality guidance (``CLI for precise operations, GUI for visual tasks''); the task guidance field carries the Path-Steer efficiency prior for this attempt (an ordered plan, ``First \dots''), used only during rollouts and empty at evaluation. The interaction history is abstracted with placeholders: each turn supplies the instruction, running action/CLI-result histories, and the current screenshot, and the agent replies with a thought and one GUI or CLI block until it terminates. Task text and coordinates are elided.}
\label{fig:rollout-message}
\end{figure*}

\begin{figure*}[t]
\centering
\begin{lstlisting}[style=prompt]
[SYSTEM]
You are a task completion evaluator for a GUI automation agent operating a desktop application.

You will be given:
1. A task instruction describing what the agent should accomplish
2. The execution trajectory showing actions taken and their results
3. Screenshots from the execution in chronological order

Your job is to judge whether the task was completed successfully.

How to read the screenshots:
- The first screenshot is only the earliest available visual baseline. Do not treat it as the result.
- Middle screenshots are context for how the state changed over time.
- The final screenshot is the primary visual evidence for task completion.
- If the trajectory says an edit/export succeeded but the final screenshot or final artifact evidence does not
  show it, score conservatively.
- If the final screenshot is blank, stale, still in a menu/dialog, or appears unchanged from the baseline, do
  not give high credit for visual tasks.
- For file-based editing tasks where the instruction says the exported file/project is the scored artifact,
  treat successful CLI output and exported artifact evidence as authoritative for file contents. Desktop
  application screenshots may be stale because many apps do not auto-refresh after external file edits; do not
  penalize missing GUI refresh when the trajectory provides structured CLI evidence that the exported artifact
  contains the requested edits.

Scoring guidelines:
- 1.0: Task fully completed, all requirements met, visual confirmation matches expectations
- 0.7-0.9: Task mostly completed, minor issues (e.g. slightly wrong values, visual looks close)
- 0.4-0.6: Task partially completed (some steps done, others missing or wrong)
- 0.1-0.3: Task barely started or mostly failed
- 0.0: Task not completed at all, or agent timed out without meaningful progress

When evaluating, consider:
- Did CLI commands succeed (rc=0) and produce expected output?
- Does the final screenshot show the expected visual result?
- Compared with the baseline screenshot, are the requested changes visible in the final screenshot?
- Were all sub-tasks in the instruction addressed?
- Did the agent reach a terminal state (DONE) or time out?

Respond with ONLY a JSON object (no markdown, no extra text):
{"score": <float 0.0-1.0>, "reason": "<brief explanation of what was/wasn't completed>"}

[USER]
# Task Instruction
{instruction}

# Execution Trajectory
{trajectory_evidence}

# Screenshots
The following {n} screenshot(s) are sampled from the trajectory and are ordered from earliest to latest. Use
the earliest screenshot only as baseline/context. The final 3 screenshots, when present, show the end-state
context. Use the FINAL screenshot as the main visual evidence for scoring.
{image_1} ... {image_n}   # base64 PNG image inputs, each labeled with its chronological role
\end{lstlisting}
\caption{Full VLM judge prompt. Braces denote per-attempt inputs: \texttt{\{instruction\}} the task instruction,
\texttt{\{trajectory\_evidence\}} the compacted action/CLI trace, and \texttt{\{image\_1..n\}} the sampled,
role-labeled screenshots supplied as image inputs.}
\label{fig:judge-prompt}
\end{figure*}

\begin{figure*}[t]
\centering
\begin{lstlisting}[style=prompt]
You are a multi-purpose intelligent assistant. Based on my requests, you can use tools to help me complete various tasks.
# Tools
You have access to the following functions:
<tools>
{computer_use_function_schema}
{execute_cli_function_schema}
</tools>
If you choose to call a function ONLY reply in the following format with NO suffix:
<tool_call>
<function=example_function_name>
<parameter=example_parameter_1>
value_1
</parameter>
<parameter=example_parameter_2>
This is the value for the second parameter
that can span
multiple lines
</parameter>
</function>
</tool_call>
<IMPORTANT>
Reminder:
- Function calls MUST follow the specified format: an inner <function=...></function> block must be nested within <tool_call></tool_call> XML tags
- Required parameters MUST be specified
- You may provide optional reasoning for your function call in natural language BEFORE the function call, but NOT after
- If there is no function call available, answer the question like normal with your current knowledge and do not tell the user about function calls
- The current date is {runtime_date}.
- Collapsed screenshots appear as text: This screenshot has been collapsed.
</IMPORTANT>
# Response format
Response format for every step:
1) Action: a short imperative describing what to do.
2) One or more <tool_call>...</tool_call> blocks, each containing one tool call.
Rules:
- Output exactly in the order: Action, then the <tool_call> block(s).
- If multiple <tool_call> blocks are needed, they will be executed sequentially in the order you output them.
- Prefer one <tool_call> for normal steps; use multiple only for tightly coupled low-level actions such as click-then-type or key sequences.
- Use <function=execute_cli> for CLI commands. Use <function=computer_use> only for GUI actions.
- Do not mix execute_cli with other tool calls in the same step unless the command is immediately required by the same atomic action.
- Be brief: one sentence for Action.
- Do not output anything else outside those parts.
- If finishing, use action=terminate in the tool call.
# CLI mode rules
- This evaluation is in CLI mode, meaning the separate execute_cli function is available in addition to normal GUI actions.
- For file-based inspection, edits, saves, exports, or verification covered by the listed CLI tools, call the separate <function=execute_cli> tool before GUI interaction.
- Use GUI actions for visual editing/selection when they are more natural or when no listed CLI tool covers the operation.
- Do not use application menus, file pickers, or a different application for an operation that a listed CLI command can perform directly.
- Do not run registry tool names as shell commands. Use the concrete command form shown in the command parameter description.
- After a successful CLI edit/save/export/verification, use that CLI feedback to decide whether to terminate instead of repeating GUI confirmation loops.
- Use wait when the environment needs time after either GUI or CLI actions.
- Use terminate only after the requested result is complete.
- Never output placeholder coordinates such as [0, 0].
\end{lstlisting}
\caption{Representative evaluation actor system prompt (Qwen, CLI+GUI); each backbone has its own prompt and a GUI-mode counterpart following the OSWorld protocol. It is application-agnostic; per-application capability enters only through the task-specific command registry injected into the \texttt{execute\_cli} schema.}

\label{fig:actor-prompt}
\end{figure*}

\end{document}